\documentclass[acmsmall,screen]{acmart}

\usepackage{bbm}

\usepackage{color}
\definecolor{mgreen}{RGB}{0,152,70}
\definecolor{mred}{RGB}{162,64,54}
\definecolor{myellow}{RGB}{216,160,3}
\definecolor{mpurple}{RGB}{68,9,225}

\definecolor{myellow2}{RGB}{248,180,124}
\definecolor{mgreen2}{RGB}{95,228,182}

\usepackage{hyperref}
\usepackage{booktabs}
 \usepackage{multirow}
\usepackage{amsmath}

\usepackage{url}

\usepackage{arydshln}
\usepackage{enumerate}
\usepackage{amsfonts}
\usepackage{bm}

\newcommand{\paratitle}[1]{\vspace{0.8ex}\noindent \textbf{#1}}

\AtBeginDocument{%
  \providecommand\BibTeX{{%
    \normalfont B\kern-0.5em{\scshape i\kern-0.25em b}\kern-0.8em\TeX}}}

\setcopyright{acmcopyright}
\copyrightyear{2022}
\acmYear{2022}
\acmDOI{10.1145/3564281}

\acmJournal{TOIS}
\acmVolume{41}
\acmNumber{50}
\acmArticle{1}
\acmMonth{4}

\begin{document}

\title{On the Robustness of Aspect-based Sentiment Analysis: Rethinking Model, Data and Training}

\author{Hao Fei}
\email{haofei37@nus.edu.sg}
\author{Tat-Seng Chua}
\email{chuats@comp.nus.edu.sg}
\affiliation{%
  \institution{National University of Singapore}
  \department{Sea-NExT Joint Lab, School of Computing}
  \streetaddress{5 Prince George's Park}
  \country{Singapore}
  \postcode{118404}
}

\author{Chenliang Li}
\email{cllee@whu.edu.cn}
\author{Donghong Ji}
\email{dhji@whu.edu.cn}
\affiliation{%
  \institution{Wuhan University}
  \department{Key Laboratory of Aerospace Information Security and Trusted Computing, Ministry of Education, School of Cyber Science and Engineering}
  \streetaddress{299 Bayi Road, Wuchang District}
  \city{Wuhan}
  \country{China}
  \postcode{430072}
}

\author{Meishan Zhang}
\affiliation{%
  \department{Institute of Computing and Intelligence}
  \institution{Harbin Institute of Technology (Shenzhen)}
  \streetaddress{Taoyuan St, Nanshan District}
  \city{Shenzhen}
  \postcode{518055}
  \country{China}}
\email{mason.zms@gmail.com}

\author{Yafeng Ren}
\affiliation{%
  \institution{Guangdong University of Foreign Studies}
  \department{School of Interpreting and Translation Studies}
  \streetaddress{2 Baiyun Avenue North, Baiyun District}
  \city{Guangzhou}
  \postcode{510515}
  \country{China}}
\email{renyafeng@whu.edu.cn}

\thanks{
This work is supported by the Sea-NExT Joint Lab,
the Key Project of State Language Commission of China (No. ZDI135-112), 
the National Key Research and Development Program of China (No. 2017YFC1200500), 
the Science of Technology Project of GuangZhou (No. 20210202607), 
the National Natural Science Foundation of China (No. 61772378, No. 62176187, No. 62176180), 
the Research Foundation of Ministry of Education of China (No. 18JZD015).
}

\renewcommand{\shortauthors}{Fei, et al.}

\begin{abstract}
Aspect-based sentiment analysis (ABSA) aims at automatically inferring the specific sentiment polarities towards certain aspects of products or services behind the social media texts or reviews, which has been a fundamental application to the real-world society.
Within recent decade, ABSA has achieved extraordinarily high accuracy with various deep neural models.
However, existing ABSA models with strong in-house performances may fail to generalize to some challenging cases where the contexts are variable, i.e., being low robustness to real-world environment.
In this study, we propose to enhance the ABSA robustness by systematically rethinking the bottlenecks from all possible angles, including model, data and training.
First, we strengthen the current best-robust syntax-aware models by further incorporating the rich external syntactic dependencies and the labels with aspect simultaneously with a universal-syntax graph convolutional network.
In the corpus perspective, we propose to automatically induce high-quality synthetic training data with various types, allowing models to learn sufficient inductive bias for better robustness.
Lastly, we based on the rich pseudo data perform adversarial training to enhance the resistance to the context perturbation, and meanwhile employ contrastive learning to reinforce the representations of instances with contrastive sentiments.
Extensive robustness evaluations are conducted.
The results demonstrate that our enhanced syntax-aware model achieves better robustness performances than all the state-of-the-art baselines.
By additionally incorporating our synthetic corpus, the robust testing results are pushed with around 10\% accuracy, which are then further improved by installing the advanced training strategies.
In-depth analyses are presented for revealing the factors influencing the ABSA robustness.
\end{abstract}




\maketitle

\section{Introduction}
\label{sec:introduction}

Sentiment analysis, mining the user's opinion behind the social media or product review texts, has long been a hot research topic in the communities of data mining and natural language processing (NLP) \cite{mullen-collier-2004-sentiment,wilson-etal-2005-recognizing,SchoutenF16,WangPDX17,FeiZRJ20}.
The aspect-based sentiment analysis (ABSA), aka. fine-grained sentiment analysis, as a later emerged research direction aiming to infer the sentiment polarities towards a specific aspect in text, has gained an overwhelming number of research efforts during the last decade \cite{tang-etal-2016-aspect,ClercqLJCH17,ZimbraAZC18,do2019deep,LiuS20a,LiuZG21}.
Within recent years, ABSA has secured prominent performance gains \cite{xue-li-2018-aspect,jiang-etal-2019-challenge,tang-etal-2020-dependency,chen-etal-2020-inducing,ChivukulaL19,WangCALPL21}, with the establishment of various deep neural networks.

\begin{figure}[!t]
\centering
\includegraphics[width=1.0\columnwidth]{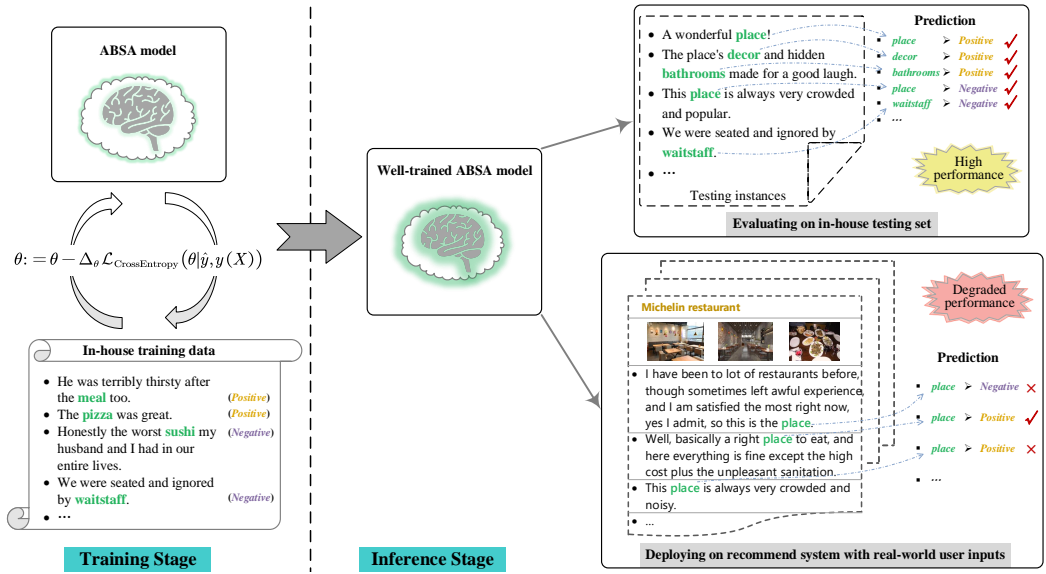}
\caption{
An example to illustrate the performance gap between in-house evaluation and real-world scenario of aspect-based sentiment analysis model.
}
\label{introduction-motivation}
\end{figure}

Although current strong-performing ABSA models have achieved high accuracy on standard test sets (e.g.,  SemEval data \cite{pontiki-etal-2014-semeval,pontiki2015semeval}), they may fail to generalize correctly to new cases in the wild where the contexts can be varying.
Especially in real-world applications, different from the enclosure test\footnote{
`Enclosure test' is to describe a process that the data used for training and testing are draw from the same sources and in same distributions.
}, ABSA system will receive all kinds of diversified inputs from a variety of users, which naturally calls for more robust ABSA models.
In Fig. \ref{introduction-motivation} we give a running example.
An ABSA system being well-trained on in-house training data can perform well when evaluated on in-house testing data.\footnote{
Here `well-trained' is used to describe an ABSA model that is trained on the in-house training set and achieves the peak performance on the developing set.
}
Unfortunately, once deploying the ABSA model on recommend system with real-world user inputs, it fails to generalize to those unseen cases in the wild, i.e., being low robustness to factual environment.

In the recent research on robustness, it was shown that the performances of current ABSA methods can drop drastically by over 50\% in terms of predicting accuracy \cite{xing-etal-2020-tasty}.
Within one piece of text, only a small subset of contexts will genuinely trigger the sentiment polarity of the target aspect (generally are opinion terms).
And correspondingly, a robust ABSA system needs to place most focus on such critical cues instead of the other trivial or even misleading clues\footnote{
We use `trivial contexts' and `misleading clues' to describe those contexts that are not the opinion expressions for triggering the aspects in ABSA.
}, and should not be disturbed by the change of non-critical background contexts \cite{xing-etal-2020-tasty}.

Based on recent study \cite{jiang-etal-2019-challenge,xing-etal-2020-tasty}, two major robustness test\footnote{
`Robustness test' is referred to as a probing test on a model in terms of its robustness, which often performed based on a certain robust testing data set.
} challenges in ABSA can be summarized, as shown in Table \ref{intro}.
The first type, \textbf{\em aspect-context binding} challenge,
requires that the target aspect should be correctly bound to its corresponding key clues, instead of some other trivial words. 
Take as the example of [Raw \emph{S\scriptsize{1}}],
altering the crucial opinion expressions (i.e., from `fabulous' to `awful') of the target aspect should directly flip its polarity (i.e., from positive into negative) as in [Mod \emph{S\scriptsize{1-1}}].
Also, diversifying the non-critical words (i.e., by altering the background trivial contexts) should not influence its polarity, as in [Mod \emph{S\scriptsize{1-2}}].
Low-robust model would be vulnerable when facing with such changes. 
Another type is \textbf{\em multi-aspect anti-interference} challenge.
When multiple aspects coexist in one sentence, the sentiment of target aspect should not be interfered by other aspects.
For example, based on [Raw \emph{S\scriptsize{2}}], adding additional non-target aspect (as in [Mod \emph{S\scriptsize{2-1}}]), especially with opposite polarity (as in [Mod \emph{S\scriptsize{2-2}}]), should not influence the judgement of the polarity for target aspect.

In this study, we explore the enhancement for ABSA robustness.
We cast the following researching questions.

\begin{itemize}
    \item[\textbf{Q1}:] \emph{What types of neural models are more robust for ABSA?}
    \item[\textbf{Q2}:] \emph{Is the current ABSA corpus informative enough for models to learn good bias with high robustness?}
    \item[\textbf{Q3}:] \emph{Will the model become more robust via better training strategy?}
\end{itemize}

\noindent These three questions together reflect the bottlenecks of ABSA robustness from different aspects, i.e., model, data and training.

With respect to the robust ABSA model, the key is to effectively model the relationship between the target aspect and its valid contexts\footnote{
`Valid contexts' describes the parts of context are critical cues of the opinion expressions.
},
e.g., using attention mechanisms \cite{wang-etal-2016-attention,tang-etal-2016-aspect,HuangOC18} or position encoding \cite{tang-etal-2016-effective}  to enhance the sense of the location of the target aspect.
Especially, a large proportion of work has shown that the leverage of syntactic dependency information helps the most \cite{zhang-etal-2019-aspect,huang-carley-2019-syntax,wang-etal-2020-relational,xing-etal-2020-tasty}.
We however note that most existing syntax-aware models only integrate the word dependence while leaving the syntactic dependency labels unemployed.
Actually the dependency arcs with different types carrying distinct evidences may contribute in different degrees, which may help to better infer the relations between aspect and valid clues.
Thus, how to better navigate the rich external syntax for better robustness still remains unexplored.

\begin{table}[!t]
\caption{Challenges of robustness tests in aspect-based sentiment analysis.
}
% \vspace{-12pt}
\begin{center}
\resizebox{1.0\columnwidth}{!}{
  \begin{tabular}{llr}
\hline\hline
\multicolumn{3}{l}{$\blacktriangleright$Challenge\#1. {\textbf{\em Aspect-context binding}}} \\
\multicolumn{3}{l}{\qquad The target aspect correctly bounds to its corresponding key clues, instead of the other trivial or wrong contexts.}\\
\hdashline
\multicolumn{3}{l}{Examples: }\\
\quad [Raw \emph{S\scriptsize{1}}] & The \textcolor{mgreen}{\textbf{food}} is fabulous, and anyway we enjoy the journey. & (\textcolor{myellow}{\textbf{\em Positive}}) \\
\quad [Mod \emph{S\scriptsize{1-1}}] & The \textcolor{mgreen}{\textbf{food}} is \underline{awful}, and anyway we enjoy the journey. &(\textcolor{mpurple}{\textbf{\em Negative}})\\
\quad [Mod \emph{S\scriptsize{1-1}}] & The \textcolor{mgreen}{\textbf{food}} \underline{tastes} fabulous, and \underline{overall I} enjoy the journey. & (\textcolor{myellow}{\textbf{\em Positive}})\\
\hline\hline
\multicolumn{3}{l}{$\blacktriangleright$Challenge\#2. {\textbf{\em Multi-aspect anti-interference}}} \\
\multicolumn{3}{l}{\qquad Multiple aspects coexist and the sentiment of target aspect should not be interfered by other aspects.}\\
\hdashline
\multicolumn{3}{l}{Examples:} \\
\quad [Raw \emph{S\scriptsize{2}}] & The \textcolor{mgreen}{\textbf{seafoods}} here are the best ever. &(\textcolor{myellow}{\textbf{\em Positive}}) \\
\quad [Mod \emph{S\scriptsize{2-1}}] & The \textcolor{mgreen}{\textbf{seafoods}} here are the best ever, as well as the \emph{attentive} \textcolor{mgreen}{\textbf{service}}. &(\textcolor{myellow}{\textbf{\em Positive}})\\
\quad [Mod \emph{S\scriptsize{2-1}}] & The \textcolor{mgreen}{\textbf{seafoods}} here are the best ever, except the \emph{noisy} \textcolor{mred}{\textbf{ambient}}. &(\textcolor{myellow}{\textbf{\em Positive}})\\
\hline\hline
\end{tabular}
}
\end{center}
  \label{intro}
\end{table}

As for corpus, almost all the ABSA models are trained and evaluated in an enclosure based on the SemEval dataset \cite{pontiki-etal-2014-semeval,pontiki2015semeval,pontiki-etal-2016-semeval}.
But 80\% sentences in these datasets have either single aspect or multiple aspects in same polarity \cite{jiang-etal-2019-challenge}.
Even a well-trained ABSA model on such data will suffer performance downgrading when exposed to an open environment with complex inputs.
Ideally, a training corpus with varying and challenging instances would enable ABSA models to be more robust.
Yet manually annotating data is labor-intensive, which makes automatic high-quality data acquisition indispensable.
Besides, most of the ABSA frameworks are optimized directly towards the gold targets with cross-entropy loss.
This will inevitably lead to low-efficient utilization of training data, or even poor resistance to perturbation.
We believe a better training strategy could help to excavate the knowledge behind the data more efficiently and sufficiently.

To this end, we aim to enhance the ABSA robustness by rethinking the model, data and training, respectively, for each of which we propose retrofitting solution.
First, we introduce a universal-syntax graph convolutional network (namely, USGCN) for incorporating the syntactic dependencies with labels simultaneously.
By effectively modeling rich syntactic indications, USGCN learns to better reason between aspect and contexts.
Second, we present an algorithm for automatic synthetic data construction.
Three types of high-quality pseudo corpora are induced based on raw data, enriching the data diversification for robust learning.
Third, we leverage two enhanced training strategies for robust ABSA, including the adversarial training \cite{ebrahimi-etal-2018-hotflip,ribeiro-etal-2018-semantically}, and the contrastive learning \cite{He0WXG20,ChenK0H20}.
Based on the synthetic training data, the adversarial training help to reinforce the perception of contextual change,
while the contrastive learning further unsupervisedly consolidates the recognition of different labels.

\begin{figure}[!t]
\centering
\includegraphics[width=0.98\columnwidth]{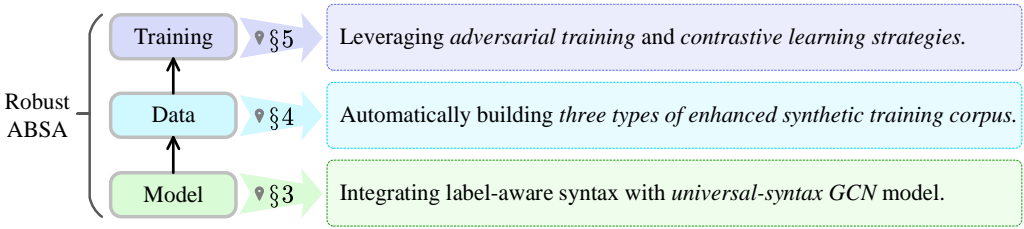}
\caption{
A high-level overview of our solutions for enhancing the robustness of aspect-based sentiment analysis.
}
\label{procedure}
\end{figure}

We note that the two main challenges of ABSA as shown in Table \ref{intro} can be essentially the same, i.e., they are the two sides of a coin. 
In this paper, all the three perspectives (i.e., model, data and training) and the corresponding methods we proposed here all target solving these two challenges. 
For example, we propose three different synthetic data construction methods, where the sentiment modification method ($\S$\ref{Sentiment Modification}) and the background rewriting method ($\S$\ref{Background Rewriting}) both directly target addressing the \textbf{\em aspect-context binding} challenge, and the non-target aspects addition method ($\S$\ref{Non-target Aspects Addition}) is proposed for relieving the \textbf{\em multi-aspect anti-interference challenge}. 
With respect to the model, our proposed syntax-aware ABSA model ($\S$\ref{Syntax-aware Neural Model}) can enhance the \textbf{\em aspect-opinion binding}, which essentially indirectly solves the \textbf{\em aspect anti-interference} issue. 
And the advanced training strategies ($\S$\ref{Towards Robustness Training}) help solve both the two challenges.
We neatly summarize the overall proposal in Fig. \ref{procedure}.

We perform extensive experiments on multiple robustness testing datasets for ABSA.
Experimental results show that the USGCN model achieves the most robust performances than all the state-of-the-art baseline systems.
All the ABSA models substantially achieve enhanced robustness when additionally using our pseudo training data, which can be further strengthened by installing the advanced training strategies.
Further in-depth analyses from multiple aspects have been conducted for revealing the factors influencing the ABSA robustness.

In general, the contributions of our work are as follows.

\begin{itemize}
    \item[1).] We propose a novel syntax-aware model: we model the syntactic dependency structure and the arc labels as well as the target aspect simultaneously with a GCN encoder, namely universal-syntax GCN (USGCN). With USGCN, we achieve the goal of navigating richer syntax information for best ABSA robustness.
    \item[2).] We build an algorithm for automatically inducing high-quality synthetic training data with various types, allowing models to learn sufficient inductive bias for better robustness. Each type of pseudo data aims to improve one certain angle of ABSA robustness.
    \item[3).] We perform adversarial training based on the pseudo data to enhance the resistance to the environment perturbation. Meanwhile, we employ the unsupervised contrastive learning technique for further enhancement of representation learning, based on the contrastive samples in pseudo data.
    \item[4).] Our overall framework has achieved significant improvements on robustness test on the benchmark datasets. In-depth analyses have been presented for revealing the factors influencing the ABSA robustness.
\end{itemize}

The remainder of the article is organized as follows.
Section \ref{Related Work} surveys the related work.
Section \ref{Syntax-aware Neural Model} elaborates in detail the enhanced syntactic-aware ABSA neural model.
In Section \ref{Synthetic Corpus Construction}, we present the algorithm for the synthetic training corpus construction.
Section \ref{Towards Robustness Training} shows how to perform the advanced training strategies.
Section \ref{Experiment} gives the experimental setups and the results on the robustness study of our system.
Section \ref{Analysis and Discussion} analyzes in deep the factors influencing the ABSA robustness.
Finally, in Section \ref{Conclusion and Future Work} we present the conclusions and future work.

\section{Related Work}
\label{Related Work}

In this section, we give a literature review of related work on sentiment analysis and the robustness study of aspect-based sentiment analysis.

\subsection{Sentiment Analysis and Opinion Mining}

Sentiment analysis or opinion mining aims to use machines to automatically infer the sentiment intensities or attitudes of the texts generated by users in the Internet \cite{PangL07,PhienthrakulKTO09,dong-etal-2014-adaptive,0001LRZJ22}.
Since it shows great impacts to the real-world society, sentiment analysis facilitates a wide range of downstream applications, and has long been fundamental research direction in the community of NLP and data mining within the past decades \cite{RenZZJ16,WangPDX17,LiuCWMZ18}. 
Initial methods for sentiment analysis employ the rule-based models, e.g., using sentiment or opinion lexicons, or designing hard-coded regular expressions \cite{PangL07,PhienthrakulKTO09}.
Then, researchers incorporate statistical machine learning models with hand-crafted features for the tasks \cite{mullen-collier-2004-sentiment,wilson-etal-2005-recognizing}.

In the last decade, the deep learning methods have received great attention.
Neural networks together with continuous distributed features are extensively adopted for enhancing the task performances of sentiment analysis \cite{RenZZJ16,li-etal-2018-transformation,0001RWLJ21,9634849}.
In particular, the Long-short Term Memory (LSTM) models \cite{hochreiter1997long}, Convolutional Neural Networks (CNN) \cite{Kim14}, Attention mechanisms \cite{wang-etal-2016-attention,weston2014memory}, Graph Convolutional Networks (GCN) \cite{zhang-etal-2019-aspect,Wu0RJL21} are the most notable deep learning methods that have been extensively adopted for sentiment analysis.
For example, Wang et al. (2016) \cite{wang-etal-2016-attention} propose an attention-based LSTM network for attending different parts of the aspects for aspect-level sentiment classification.
Xue et al. (2018) \cite{xue-li-2018-aspect} propose CNN-based model with gating mechanisms for selectively learning the sentiment features meanwhile keeping computational efficient with convolutions.
Zhang et al. (2019) \cite{zhang-etal-2019-aspect} build a GCN encoder
over the syntactic dependency trees of sentences to exploit syntactical information and word dependencies.

On the other hand, the research focus has been shifted into ABSA that detects the sentiment polarities towards the specific aspects in the sentence \cite{VargheseJ13,pontiki-etal-2014-semeval}.
Compared with the standard coarse-grained (i.e., sentence-level) sentiment analysis, such fine-grained analysis shows more impacts on the real-world scenario, such as social media texts and product reviews, and thus facilitate a wider range of downstream applications.
Prior methods for sentiment analysis mostly employ statistical machine learning models with manually-crafted discrete features \cite{mullen-collier-2004-sentiment,wilson-etal-2005-recognizing,PhienthrakulKTO09}.
Later, neural networks together with continuous distributed features, as used in sentence-level sentiment analysis, are extensively adopted to achieve big wins \cite{dong-etal-2014-adaptive,jiang-etal-2019-challenge,zhang-etal-2019-aspect,huang-carley-2019-syntax}.
The difference of the neural models between coarse-grained sentiment analysis and ABSA lies in that, the ABSA needs additionally to model the target aspect concerning its contexts.
Tang et al. (2016)  \cite{tang-etal-2016-aspect} use a memory network to cache the sentential representations into external memory and then calculate the attention with the target aspect.
Recently, Veyseh et al. (2020) \cite{pouran-ben-veyseh-etal-2020-improving} regulate the GCN-based representation vectors based on the dependency trees in order to benefit from the overall contextual importance scores of the words.

\subsection{Robustness Study of Aspect-based Sentiment Analysis}

Analyzing the robustness of learning systems is a crucial step prior to models' deployment.
Robustness study thus has been an important research direction in many areas.
A highly-performing system on test set may fail to generalize to new examples where the contexts are varying, such as with distribution shift or adversarial noises \cite{jia-liang-2017-adversarial,iclr-BelinkovB18,MillerKRS20}.
Similarly, given the fact that current state-of-the-art ABSA models obtain high scores on the test datasets, they could be low in robustness.
In recent ABSA robustness probing study \cite{jiang-etal-2019-challenge,xing-etal-2020-tasty}, all of the existing models show huge accuracy degradation when testing on the robustness test set.
It thus becomes imperative to strength the ABSA robustness.
However, unlike the robustness problem in other NLP tasks such as text classification, ABSA task characterizes that multiple aspect mentions and their supporting clues can be intertwined together in one sentence, which make it more difficult to solve.
As we argue earlier, there are at least three angles to begin with, i.e., model, data and training.

Various neural models are investigated for better ABSA, e.g., RNN \cite{tang-etal-2016-effective}, memory networks \cite{tang-etal-2016-aspect}, attention networks \cite{wang-etal-2016-attention,HuangOC18}, graph networks \cite{huang-carley-2019-syntax,wang-etal-2020-relational,tang-etal-2020-dependency}, etc.
Later research has repeatedly shown that the syntactic dependency trees are of great effectiveness for ABSA, since such information provide additional signals to help to infer the relations between target and valid contexts \cite{zhang-etal-2019-aspect,huang-carley-2019-syntax,wang-etal-2020-relational}.
Very recent study \cite{xing-etal-2020-tasty} however show that, many of those neural models even achieving high accuracies on standard test sets, such as attention or memory mechanisms etc., are with low robustness.
They have revealed that explicit aspect-position modeling (such as syntax-aware models) and pre-trained language models show better robustness.
As we find that the arc labels in dependency structure that also can be much useful are abandoned by existing syntax-aware models.
We thus present a better solution on leveraging the external syntax knowledge, i.e., simultaneously modeling the dependency arcs and types with graph models.
Besides, we in our experiments further explore the possibility if better pre-trained language models (PLM) can improve robustness.

As preliminary works noted, most sentences in current ABSA datasets (i.e., SemEval) contain either single aspect or multiple aspects in same polarity, which downgrades the problem to coarse-grained (sentence-level) sentiment classification \cite{jiang-etal-2019-challenge,Xu19Failure}.
This underlies the weak robustness of current ABSA models which even have high performances on the testing sets.
To combat that, Jiang et al. (2019) \cite{jiang-etal-2019-challenge} newly craft a much more challenging data set, in which each sentence consists of at least two aspects with different sentiment polarities (i.e., multi-aspect multi-sentiment, MAMS data).
They show that MAMS can prevent ABSA from degenerating to sentence-level sentiment analysis and thus improve the ABSA robustness.
We also in later experiments show that training with this data enables ABSA models to be more generalizable.
However, we note that robust-driven MAMS data is fully annotated with human labor, which can cause huge cost.
To ensure the data diversification for robust learning meanwhile avoiding manual costs, we in this work consider a scalable method for automatic data construction.
We obtain three types of high-quality pseudo corpora, including 
1) flipping the sentiment of target aspect,
2) rewriting the background contexts of target aspect,
and 3) adding extra non-target aspects.
With respect to this, our work partially draws some inspirations from recent work of Xing et al. (2020) \cite{xing-etal-2020-tasty}.
Yet we differ from their work in four ways.
First, they locate the crucial opinion expressions for each target by additionally using the existing labeled TOWE data \cite{fan-etal-2019-target}, while our automatic algorithm finds such valid expressions heuristically.
Second, they only construct a small set for testing, but we construct larger volumes of data for training.
Third, they take human evaluation for quality speculation, and our method ensures high quality of data without human interference.
What's more, we consider diversifying background contexts of examples, which is fallen out of their consideration.

This work also relates to the adversarial attack training, which alters the input slightly to keep the original meaning but leads to different predictions, has been a long-standing method to enhance the robustness of NLP systems \cite{ebrahimi-etal-2018-hotflip,ribeiro-etal-2018-semantically,abs-2005-05909,abs-2009-09191}.
We note that adversarial training strategy has been employed by some existing ABSA works, but for improving the in-house performance \cite{li-etal-2019-transferable,abs-2001-11316,LiangY0DHX20}.
In this work, we for the first time design the adversarial training based on the multiple types of synthetic training data to reinforce the model perception of contextual change, so as to obtain a better environment independence.
On the other hand, we based on the synthetic examples further employ the contrastive learning algorithm to unsupervisedly consolidate the representations of examples with different polarities at high-dimension space.
Contrastive learning is a novel unsupervised or self-supervised approaches which has recently been successfully employed in multiple areas, e.g., computational vision and NLP \cite{VelickovicFHLBH19,he-etal-2020-contrastive,He0WXG20}.
The main idea is to force a model to narrow the distance between those examples with the similar target, and meanwhile widen those with different targets.
To our knowledge, we are the first utilizing contrastive learning technique on robust ABSA learning.

\begin{figure}[!t]
\centering
\includegraphics[width=0.75\columnwidth]{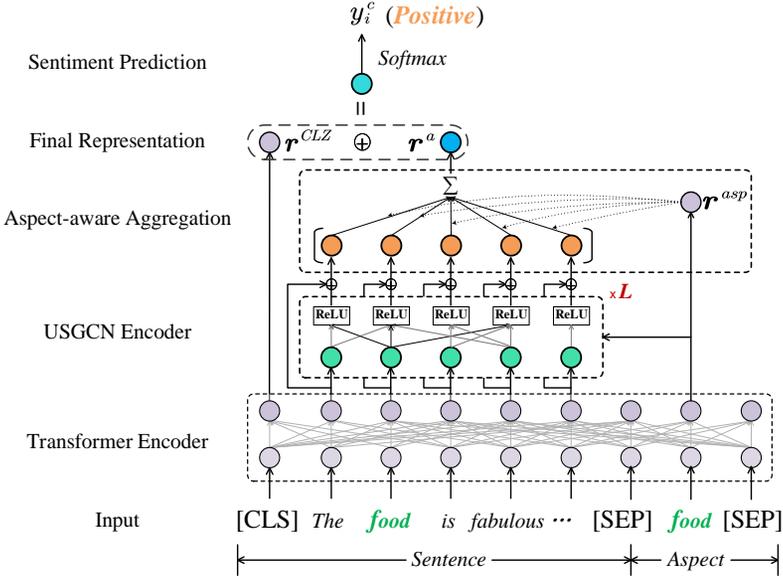}
\caption{
The overall framework for aspect-based sentiment analysis.}
\label{framework}
\end{figure}

\section{Syntax-aware Neural Model}
\label{Syntax-aware Neural Model}

\paratitle{Task Formulation.}
The goal of ABSA is to determine the sentiment polarity towards a specific aspect, which we formalize as a classification problem on
sentence-aspect pairs.
Technically, given an input sentence $X=\{x_1,\cdots,x_n\}$ and an aspect term $A=\{x_i,\cdots,x_j\}$ that is a sub-string of input sentence $X$, the model is expected to predict the corresponding sentiment label $\hat{y}$. 
Note that one sentence may contain multiple aspect terms, and we can correspondingly construct multiple sentence-aspect pairs for one sentence under a one-to-many mapping.
Through our framework the classification can be formalized as:
\begin{equation}
y^C = \mathop{\text{argmax}}_{y \in C} p(y | X, A) \,,
\end{equation}
where $C$ denotes the set of all sentiment polarity labels, i.e., `\emph{Positive}', `\emph{Negative}' and `\emph{Neutral}'.

\paratitle{Model Overview.}
The proposed neural framework mainly consists of three layers: the base encoder layer, the syntax fusion layer and the aggregation layer.
The base encoder layer employs the Transformer model \cite{VaswaniSPUJGKP17}, taking as input the sentence and the aspect term, yielding contextualized word representations as well as the aspect term representation.
The syntax fusion layer, also as our proposed universal-syntax graph convolutional network (USGCN), fuses the rich external syntactic knowledge into the feature representations.
Finally, aggregation layer summarizes and gathers the feature representations into total final one, based on which the classification layer makes prediction.
The overall framework is shown in Fig. \ref{framework}.

\begin{figure}[!t]
\centering
\includegraphics[width=1.0\columnwidth]{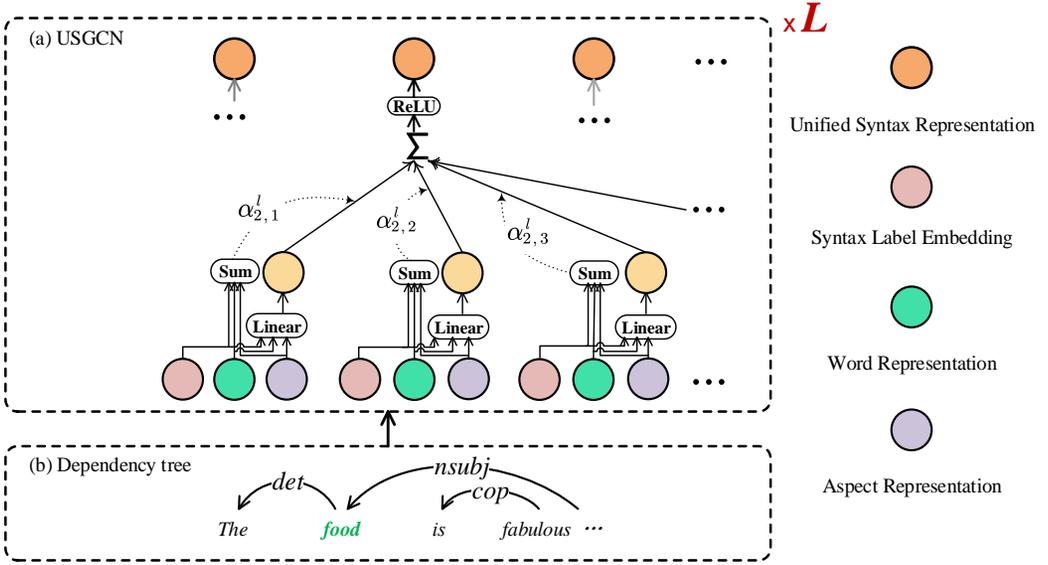}
\caption{
Illustration of the proposed (a) universal-syntax GCN (USGCN) based on the (b) syntactic dependency tree of input sentence.
}
\label{USGCN}
\end{figure}

\subsection{Base Encoder Layer}

We employ multi-layer Transformer to yield contextualized word representation $\bm{h}^X_i$ as well as the aspect word representation $\bm{r}^{asp}$.
Transformer encoder has been shown prominent on learning the interaction between each pair of input words, leading to better contextualized word representations.
Technically, in Transformer encoder, the input $\bm{x}$ is first mapped into queries $\bm{Q}$, values $\bm{V}$, and keys $\bm{K}$ via linear projection.
We then compute the relatedness between $\bm{K}$ and $\bm{Q}$ via Scaled Dot-Product alignment function, which is multiplied by values $\bm{V}$:
\begin{equation}
\label{self attention}
\bm{\alpha} = \text{Softmax}(\frac{\bm{Q}\cdot\bm{K}^{\mathrm{T}}}{\sqrt{d_{k}}}) \cdot \bm{V}
\end{equation}
where $d_{k}$ is a scaling factor.
$\bm Q$, $\bm K$ and $\bm V$ are the same of input words in our practice.
Multiple parallel attention heads focus on different parts of semantic learning.
Also, we can alternatively take the pre-trained BERT parameters \cite{devlin-etal-2019-bert} as Transformer's initiation to boost the performances.

To form an input sequence, we first concatenate the input sentence $X$ and the aspect term $A$ and combine some special tokens: $\hat{X}=\{ `CLS', X , `SEP', A, `SEP' \}$,
where `\emph{CLS}' is a symbol token for yielding sentence-level overall representation, and `\emph{SEP}' is a special token for separating the sentential words and the aspect terms.
In total, we can summarize the calculations in base encoder as follows:
\begin{equation}
 \{\bm{h}^{CLS}, \bm{H}^{X}, \bm{H}^{asp}\}  = \text{Trm}(\hat{X}) \,,
\end{equation}
where $\bm{h}^{CLS}$ is the representation for overall sentence.
% from `\emph{CLS}' token in BERT.
$\bm{H}^{X} = \{ \bm{h}^X_1, \cdots, \bm{h}^X_n\}$ are the sentential word representations, 
and $\bm{H}^{asp} = \{ \bm{{h}}^{asp}_i, \cdots, \bm{{h}}^{asp}_j\}$ are the aspect term representations which will be pooled into one $\bm{r}^{asp}$.

\subsection{Syntax Fusion Layer}

We further fuse the dependency syntax for feature enhancement.
Previous works for ABSA unfortunately merely make use of the syntactic dependency edge features (i.e., the tree structure) \cite{gao-etal-2017-implicit,huang-carley-2019-syntax,pouran-ben-veyseh-etal-2020-improving,0001RJ20}.
Without modeling the syntactic dependency labels attached to the dependency arcs, prior studies are limited by treating all word-word relations in the graph equally \cite{FeiRJ20,FeiWRLJ21,0001LLJ21,0001RJ20a}.
Intuitively, the dependency edges with different labels can reveal the relationship more informatively between target aspect and the crucial clues within context,
as exemplified in Fig. \ref{dep-tree}:

\begin{figure}[!h]
\centering
\includegraphics[width=0.58\columnwidth]{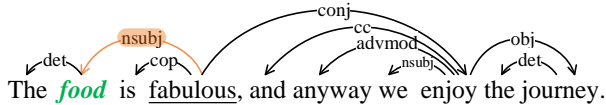}
\caption{
An example of syntactic dependency structure with edges types, based on the sentence of [Raw \emph{S\scriptsize{1}}] in Table \ref{intro}.
}
\label{dep-tree}
\end{figure}

\noindent Compared with other type of arcs within the syntax structure, the one with \emph{nsubj}\footnote{The dependency labels \emph{nsubj} refers to the nominal subject.} can present the most distinctive clues to locate the aspect `\emph{food}' with its direct opinion term `\underline{fabulous}' which strongly guides the sentiment polarity.

Also, graph convolutional network (GCN) \cite{zhang-etal-2019-aspect} has proven effective in aggregating the feature vectors within a syntactic structure of neighboring nodes, and propagating the information of a node to its neighbors.
Based on GCN, we propose a novel USGCN, modeling the dependency arcs and labels with the target aspect term simultaneously, as shown in Fig. \ref{USGCN}(a).
Technically, given the input sentence $s$ with its corresponding dependency parse (including edges $\Omega$ and labels $\Gamma$).
We define an adjacency matrix $B = \{b_{i,j}\}_{n \times n}$ for dependency edges between each pair of words, i.e., $w_i$ and $w_j$, where $b_{i,j}$=1 if there is an edge ($\in \Omega$) between $w_i$ and $w_j$, and $b_{i,j}$=0 vice versa. 
There is also a dependency label matrix $R = \{r_{i,j}\}_{T \times T} $, where each $r_{i,j}$ denotes the dependency relation label ($\in \Gamma$) between $w_i$ and $w_j$.
In addition to the pre-defined labels in $\Gamma$, we additionally add a `\emph{self}' label as the self-loop arc $r_{i,i}$ for $w_i$, and a `\emph{none}' label for representing no arc between $w_i$ and $w_j$.
We maintain the vectorial embedding $\bm{x}^e_{i,j}$ for each dependency label in $\Gamma$.

USGCN consists of $L$ layers, and 
we denote the resulting hidden representation of $w_t$ at the $l$-th layer as $\bm{r}^{l}_i$:
\begin{equation}
\bm{r}^{l}_i = \text{ReLU}(  \sum_{j=1}^n  \alpha_{i,j}^{l} ( \bm{W}_a^{l} \cdot [ \bm{r}^{l-1}_j ; \bm{x}^e_{i,j} ; \bm{r}^{asp}] + b^{l} ) ) \,,
\end{equation}
where $\bm{W}^{l}$ is parameters, $b^{l}$ is the bias term, $[;]$ denotes concatenation, and $\alpha_{i,j}^{l}$ is the neighbor connecting-strength distribution calculated via a softmax function:
\begin{equation}
\label{USGCN-attention}
\alpha_{i,j}^{l} = \frac{ b_{i,j} \cdot \exp{(\bm{W}_b^{l}[\bm{r}^{l-1}_j;\bm{x}^e_{i,j}; \bm{r}^{asp}}])}{ \sum_{k=1}^n b_{i,k} \cdot \exp{( \bm{W}_b^{l}[\bm{r}^{l-1}_k;\bm{x}^e_{i,k}; \bm{r}^{asp}])}  } \,.
\end{equation}
% where $\bm{x}^{n}_i$ is the element-wise addition by $\bm{e}^{(l-1)}_i + \bm{x}^r_{i,j}$.
% where $\bm{W}_c^{l}$ is parameter, and $[;]$ denotes the concatenating operation.
The weight distribution $\alpha_{i,j}$ entails the structural information from both the dependent edges and the corresponding labels jointly with the target aspect, and thus can comprehensively reflect the syntactic attributes towards aspect.
Note that for the first-layer USGCN, $\bm{r}^{0}_i=\bm{h}_i^X$.

\subsection{Aggregation Layer}

Next, we perform an aspect-aware aggregation to collect the salient and useful features relevant to target aspect.
\begin{equation}
\begin{aligned}
\bm{v}_{i} &= \mathrm{Tanh}( \bm{W}_c [ \bm{r}^{L}_i ; \bm{r}^{asp}] + b)  \,,\\
\beta_{i} &= \mathrm{Softmax}(\bm{v}_{i})  \,, \\
\bm{r}^{a} &= \sum \beta_{i} \cdot \bm{r}^{L}_i
\end{aligned}
\end{equation}
We then concatenate $\bm{r}^{a}$ with the sentence representation $\bm{r}^{CLS}$ into a final feature representation $\bm{r}^f$,
based on which we finally apply a softmax function for predicting ${y}^c$.

\section{Synthetic Corpus Construction}
\label{Synthetic Corpus Construction}

Synthetic data construction is a popular direction in NLP community, which effectively helps relieve the data annotation issues, such as data scarcity \cite{samanta-etal-2019-improved}, label imbalance \cite{chia-etal-2022-relationprompt} and cross-lingual data \cite{fei-etal-2020-cross}.
In this section, we elaborate the synthetic corpus construction for diversifying the raw training data (denoted as $\mathbb{D}_o$).
We mainly introduce three types of pseudo data:
1) sentiment modification of target aspect ($\mathbb{D}_a$),
2) background rewrite of target aspect ($\mathbb{D}_n$),
and 3) extra non-target aspects addition ($\mathbb{D}_m$).
These three supplementary sets provide rich signals from different angles, together helping to learn sufficient bias for better robust ABSA.
We denote the union set of the three synthetic data as $\mathbb{D}_s = \mathbb{D}_a \cup \mathbb{D}_n \cup \mathbb{D}_m$.

\subsection{Sentiment Modification}
\label{Sentiment Modification}

Modifying the sentiments of aspects is the primary operation.
For $k$-th aspect $A_{i,k}$ (with polarity label $y_{i,k}^{C}$) in $i$-th original sample $X^o_i$, ($X^o_i \in \mathbb{D}_o$), we aim to generate a batch of new sentences $X_{i,k(j)}^o \in \mathbb{D}_a$ where the sentiment polarity of $A_{i,k}$ will be 
1) kept same as $y_{i,k}^{C}$,
and 2) flipped into two other labels, i.e., $y_{i,k}^{C} \mapsto y_{i,k}^{C^{'}}$.
The creation of $\mathbb{D}_a$ involves two steps: locating opinion, and changing sentiment.

\subsubsection{Locating Opinion.}
The key to sentiment modification is to locate the exact opinion texts $O_{i,k}$ of the target aspect $A_{i,k}$.
In Xing et al. (2020) \cite{xing-etal-2020-tasty}, the TOWE data \cite{fan-etal-2019-target} is used where such opinion expressions are labeled explicitly based on the SemEval data.
However, in this work we do not consider using TOWE, for multiple reasons.
First of all, TOWE is from fully manual annotation, while we aim to build a completely automatic algorithm.
Second, training ABSA models using additional labeled opinion signals (i.e., TOWE) can lead to unfair comparisons.
Instead we thus reach the goal heuristically by defining some rules.
We extract aspect's explicit opinion expressions that satisfy following syntactic dependent relations.

\begin{itemize}
    \item[1).] \textbf{\emph{amod}} (adjectival modifier) relation, for example the aspect-opinion pair ``\emph{price}''-``\emph{reasonable}'' in ``\emph{a reasonable price}''.
    \item[2).] \textbf{\emph{nsubj}} (nominal subject) relation, e.g., a pair ``\emph{room}''-``\emph{small}'' in ``\emph{the room is small}''.
    \item[3).] \textbf{\emph{dobj}} (direct object) relation, e.g., ``\emph{smell}''-``\emph{love}'' in ``\emph{I love the smell}''.
    \item[4).] \textbf{\emph{xcomp}} (open clausal complement) relation, e.g., ``\emph{beer}''-``\emph{spicy}'' in ``\emph{The beer tastes spicy}''.
\end{itemize}

\subsubsection{Changing Sentiment.}
Then, we consult the sentiment lexicon resource for opinion word replacement,
such as SentiWordNet \cite{haccianella20103}.
For example of the word `\emph{difficult}', we can obtain its antonymous opinion words ``\emph{easy}'', ``\emph{simple}'', and synonymous word ``\emph{hard}'', ``\emph{tough}'' etc.
Besides, we can flip the polarity with some negation words or adverbs.
By this we obtain a set of target candidate $O^t_{i,k(j)}$ for the replacement of source opinion $O^s_{i,k}$.
We perform such replacement one by one to get the new sentences $X_{i,k(j)}^o$.

To control the induction quality, we define a \emph{modification confidence} as the likelihood for a successful modification,
i.e., correctly finding the opinion statement \& amending the sentiment into target.
Note that with lexicon resource, for each word we can easily obtain its sentiment strength score $a(O,C) \in [0,1]$ towards three respective polarities.
For the source opinion expression $O^s_{i,k}$ we take its sentiment score $a(O^s,C_s)$ towards the source gold polarity $y_{i,k}^{C_s}$ as the opinion localization confidence.
Likewise, for $O^s_{i,k}$'s $j$-th candidate replacement $O^t_{i,k(j)}$, we collect its all three sentiment scores.
We then define \emph{modification confidence} as:
\begin{equation}%\small
    p_a(O^{s\mapsto t},y^{C_s\mapsto C_t}) = a(O^s,C_s) \cdot  \frac{2 a(O^t,C_t)}{\sum_{C_e\ne C_t} a(O^t,C_e)}    \,,
\end{equation}
where the first term $a(O^s,C_s)$ indicates the confidence of the correct opinion localization,
and the latter part $\frac{2 a(O^t,C_t)}{\sum_{C_e\ne C_t} a(O^t,C_e)}$ indicates the sentiment flipping confidence.
We filter out those cases whose \emph{modification confidence} is lower than a pre-defined threshold $\theta_a$.

There are also several special cases worth noting.
For example, we always keep the candidate opinion terms whose Part-of-Speech (POS) tags are the same as the source opinion terms.
Specifically, for those target (after modification) opinion terms with the same POS tags as source opinion terms in the original sentences, we believe there is a high alignment in between, and the opinion positioning will be accurate. 
And those candidates should be kept.
Besides, in some case, e.g., with neutral sentiment or opinions in \emph{dobj} and \emph{xcomp} syntax relations, adding negation words will be the only way for sentiment modification.
For example, to change the sentiment state of the instance ``\emph{I will try this restaurant next time.}'', the only feasible manner is to add the negation word ``'not': ``\emph{I will not try this restaurant next time}''.
Also, it is more likely to find more than one potential opinion expressions that can determine the aspect's sentiment, and we conduct modifications combinatorially.
Specifically, when multiple opinion expressions are detected, we can modify all those opinion expressions with the target replacements at the simultaneously. For each opinion expressions we perform the sentiment flipping just the same way as for the single-opinion case.

\subsection{Background Rewriting}
\label{Background Rewriting}

To enhance the robustness of ABSA models, it is important to not only diversify the opinion changes of aspects, but also enrich the background contexts.
We rewrite the non-opinion expression in original sentence $X^o_i$ of aspect terms to form new sentences $X^n_{i,k} \in \mathbb{D}_n$.

We mainly consider the following three strategies.

\begin{itemize}
    \item[1).] Changing the opinion-less contexts, such as morphology, tense, personal pronoun, punctuation and quantifier etc.
    Morphology reflects the structure of words and parts of words, e.g., stems, prefixes, and suffixes, to name a few, and transforming the words with its morphological derivation can diversify the contexts, such as ``\emph{heterogeneous}'' vs. ``\emph{homogeneous}''. 
    Also replacing the original tense or personal pronoun in a sentence with others can cater to the need.
    And adding the punctuation or changing quantifier also leads to context modification.

    \item[2).] Substituting those neutral words with its synonym or antonym\footnote{Replacing with words having same POS tags.} by looking up the WordNet \cite{Miller1995WordNet}.
    This is partially the same as the step in $\S$\ref{Synthetic Corpus Construction}, but we only make modifications for those words with neutral-opinion labels, i.e., by first consulting its sentiment intensity with SentiWordNet.

    \item[3).] Paraphrasing the original sentence via back-translation, e.g., first translating into other languages\footnote{Majority languages, including five language pairs other than English:  Chinese-English, French-English, German-English, Spanish-English and Portuguese-English.
    According to the recent findings in NMT, the performances of back-translation in those languages are satisfactory \cite{huang-etal-2021-comparison,liu-etal-2021-complementarity-pre,chiang-etal-2022-breaking}.
    } then translating them back into source language.\footnote{We employ the off-the-shelf Translation system for high-quality translation, i.e., Google Translation \url{https://translate.google.com/}.
    }
    Intuitively, the background texts of the raw sentences may be re-phrased after the back-translation but the core semantic idea is not totally changed, we reach of goal of background expression rewriting.
    Note that we also keep the target aspect term unchanged after the back-translation.
    We have following three cases. 
    1) The opinion terms are not changed during back-translation, which is the best case we desire. 
    2) The opinion terms are partially changed, i.e., being replaced as a part of the raw phrase. For example, the phrase ``French fries'' may be turned into word ``fries'', but the meaning is not changed. For such case, we will replace the translated partial expression with the original opinion expression. 
    3) The target opinion words are totally changed after the back-translation. For this case, we first consider using the sentiment lexicon SentiWordNet to find the most likely target opinion words that are the correspondence of the original opinion expression. If the likelihood is considerable, i.e., the sentiment polarity agreements are over 0.5 between the target one and the original one, we then replace the translated expression with the original opinion expression.

\end{itemize}

We maintain the validity of such modification for the rewritten sentence with the METEOR metric \cite{BanerjeeL05}, i.e., the \emph{rewriting confidence}:
\begin{equation}
\begin{aligned}
    p_n(X^n_{i,k}) &= \text{METEOR}(X^n_{i,k}) \,. \\
\end{aligned}
\end{equation}
METEOR measures the sentence on its fluency by taking into consideration the matching rationality at the whole corpus level.
We define a threshold $\theta_n$, and drop low-quality modification, i.e., $p_n(X^n_{i,k})$<$\theta_n$.

\subsection{Non-target Aspects Addition}
\label{Non-target Aspects Addition}

Finally, we add non-target aspects in existing sentences to create multiple-aspect coexistence cases.
The construction of $\mathbb{D}_m$ consists of three steps.
First, for all the aspects at the corpus level we locate the opinion-aspect expressions with the method described in $\S$\ref{Sentiment Modification}.
We then extract the minimum text unit containing the opinion-aspect expression from different sentences.
Inspired by Xing et al. (2020) \cite{xing-etal-2020-tasty}, we extract the linguistic branch (e.g., noun/verb phrases) in a constituency structure, such as ``\emph{a reasonable price}'' etc.
Second, we perform grouping on all aspects based on their embeddings derived from a pre-trained language model (e.g., BERT), so as to gather the semantic relevance score between each pair of aspect, i.e., $\phi(A,\hat{A}) \in [0,1]$.

Third, we select certain number (top $J$) of non-target aspects $\hat{A}_{i,k(j)}$ for each target aspect by the descending order of their correlation degrees.
We then concatenate the original sentence $X^o_i$ of target aspect $A_{i,k}$ with the opinion-aspect expressions of non-target aspects, as new sentence $X^m_{i,k} \in \mathbb{D}_m$.
Note that for each $X^m_{i,k}$ we keep the expressions of non-target aspects diversified on their sentiment polarities.
Also, we can construct more than one pseudo sentence for each target aspect with different non-target aspects.
To control the quality of this construction, we define an \emph{addition confidence} as the average similarity score between the target and non-target aspects in pseudo sentence:
\begin{equation}
p_m(X^m_{i,k})=\frac{1}{J}\sum_j \phi(A_{i,k},\hat{A}_{i,k(j)}) \,,
\end{equation}
% $p_m(X^m_{i,k})=\frac{1}{J}\sum_j \phi(A_{i,k},\hat{A}_{i,k(j)})$, 
with those only $p_m>\theta_m$ as valid constructions.

Also it is noteworthy that it is highly possible that the linguistically replacements or modification (i.e., data augmenting techniques) used in this section will cause the resulting sentences semantically modified or even meaningless, i.e., unnatural sentences. 
For example, the change of personal pronouns is more likely to cause the semantic altering, compared with other types of methods. 
We note that we mainly adopt these altering methods that also are commonly adopted for other tasks in NLP community: changing morphology, tense, personal pronoun, punctuation and quantifier. And thus in our practice, to best avoid generating such semantically nonsensical sentences, we use the change of personal pronouns very carefully. 
For instance, we mainly perform the pronoun change for those easy sentences having very simple and few pronouns; and for the compound sentences or sentences containing many pronouns, we only consider making changes between the third-person pronouns, i.e., ``he'' and ``she'', or do not make any change.

\begin{figure}[!t]
\centering
\includegraphics[width=.95\columnwidth]{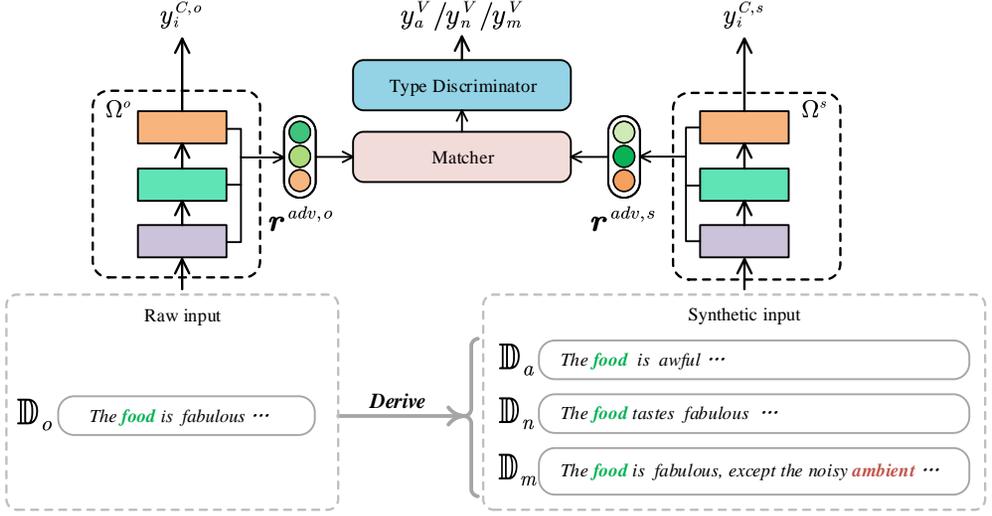}
\caption{
The adversarial training framework.
}
\label{adver-model}
% \vspace{-10pt}
\end{figure}

\section{Towards Robustness Training}
\label{Towards Robustness Training}

\paratitle{Training with Cross-entropy Objective.}
Generally, ABSA frameworks can be directly optimized towards the gold target $\hat{y}^C$ with cross-entropy objective, based on the original training data $\mathbb{D}_o$:
\begin{equation}
\label{cross-entropy}
 \mathcal{L}_e(\mathbb{D}_o) = - \frac{1}{||\mathbb{D}_o||}  \sum^{||\mathbb{D}_o||}_i \hat{y}^C_i \log y^C_i \,, %+  \frac{\zeta}{2}  ||\Theta||^{2} \,,
\end{equation}
where $||\mathbb{D}_o||$ is the length of training data.
Further based on the enriched training data, i.e., original set ($\mathbb{D}_o$) + robust synthetic corpus ($\mathbb{D}_s$ as in $\S$\ref{Synthetic Corpus Construction}), a ABSA model will achieve much better robustness with $\mathcal{L}_e(\mathbb{D}_o+\mathbb{D}_s)$.

% \begin{matrix} \end{matrix} 

\subsection{Adversarial Training}

As we mentioned earlier, ABSA models under cross-entropy training will show lower sensitivity to the environment change, leading to weak robustness.
For higher robustness, the resistance to context perturbation (e.g., opinion flip, background rewriting, and multi-aspects coexistence) should be enhanced.
We thus devise an adversarial training procedure, based on the above three kinds of synthetic training data.
As illustrated in Fig. \ref{adver-model}, in the adversarial framework, two individual neural models (as in $\S$\ref{Syntax-aware Neural Model}), $\Omega^{o}$ and $\Omega^{s}$,
1) take as input the raw sentence in $\mathbb{D}_o$ and the synthetic sentence in $\mathbb{D}_s$, respectively, 
and 2) produce middle-layer representations i.e., $\bm{r}^{adv,o}$ and $\bm{r}^{adv,s}$ respectively,
and 3) finally make their own predictions, i.e., $y^{C,o}_i$ and $y^{C,s}_i$.
Here $\bm{r}^{adv}=[\bm{r}^{CLS};\bm{r}^a;\bm{r}^s]$, where $\bm{r}^s$ is the pooled representation of $\{\bm{r}^L_1,\cdots,\bm{r}^L_n\}$ from USGCN.

The adversarial training is intermittently conducted with the regular training for $\Omega^{o}$ and $\Omega^{s}$. 
Specifically, a matcher first calculates the relatedness between $\bm{r}^{adv,o}$ and $\bm{r}^{adv,s}$:
\begin{equation}
\bm{v}  = [ \bm{r}^{adv,o} ; \bm{r}^{adv,s} ; \bm{r}^{adv,o} - \bm{r}^{adv,s} ; \bm{r}^{adv,o} \odot \bm{r}^{adv,s} ] \,, \\
\end{equation}
where the resulting representation $\bm{v}$ then is passed into the type discriminator $\mathcal{D}$ to distinguish the type of the synthetic input at $\Omega^{s}$.
We define three output of $\mathcal{D}$ as $y^V_a, y^V_n, y^V_m$ for $\mathbb{D}_a, \mathbb{D}_n, \mathbb{D}_m$, respectively.
So the adversarial goal is to achieve min-max optimization: 
minimizing the cross-entropy loss of ABSA model for the sentiment prediction $y^{C}$,
and maximizing the cross-entropy loss of the type discriminator $y^V$:
\begin{equation}
\label{adv_loss}
\begin{aligned}
    \mathcal{L}_{a_1} &= \mathop {\min }\limits_{\Omega}[ \mathop{\max} \limits_{ \mathcal{D}} ( \sum  \hat{y}^{V} \log y^{V}  )]  \,, \\
    \mathcal{L}_{a_2} &= - \sum  \hat{y}^C \log y^C \,, \\
    \mathcal{L}_{a}(\mathbb{D}_o+\mathbb{D}_s) &= \frac{1}{||\mathbb{D}_o+\mathbb{D}_s||} ( \lambda_a \mathcal{L}_{a_1}+\mathcal{L}_{a_2} ) \,,
\end{aligned}
\end{equation}
where $\lambda_a$ controls the interaction of two learning processes.
% \begin{matrix} \sum_{i=1}^D  \end{matrix}

\begin{figure}[!t]
\centering
\includegraphics[width=0.82\columnwidth]{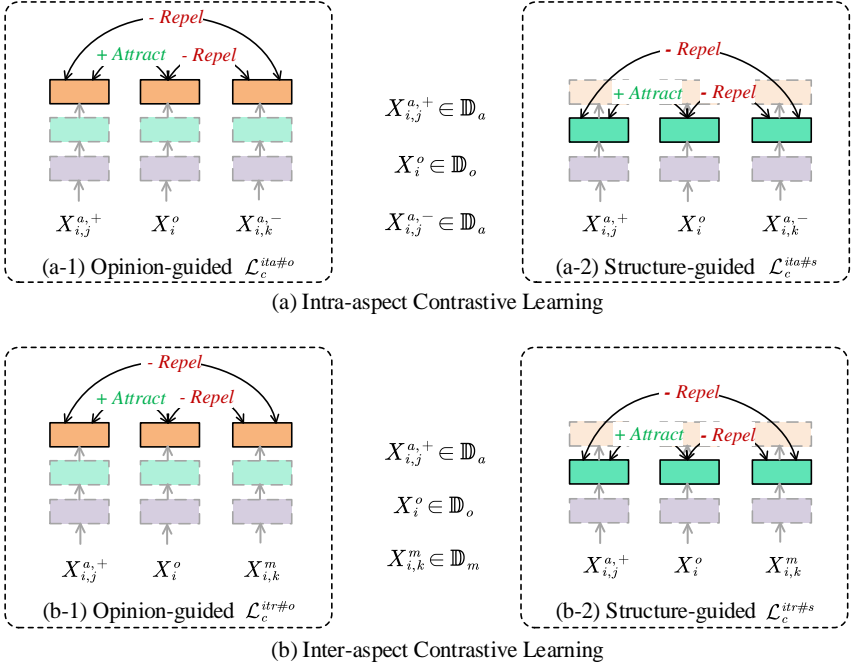}
% \vspace{-12pt}
\caption{
Different schemes of the contrastive representation learning.
The opinion-guided contrastive learning happens at aggregation layer (\textcolor{myellow2}{$\blacksquare$}), while the structure-guided contrastive learning happens at syntax fusion layer (\textcolor{mgreen2}{$\blacksquare$}).
}
\label{CLs}
% \vspace{-10pt}
\end{figure}

\subsection{Contrastive Learning}
\label{Contrastive Learning}

To sufficiently utilize the synthetic training corpus, we further employ the contrastive learning technique such that we can further consolidate the recognition of the ABSA model on different labels.
Contrastive learning has been shown effective on the representation enhancement in unsupervised manner \cite{ChenK0H20,TschannenDRGL20,TianKI20,0001WRZ22}. 
It encourages to narrow the distance between the embedding representation of examples with the similar target, and meanwhile widen those with different targets.
Here we set the goal as to increase the awareness of the ABSA model on the sentiment changes of target aspects caused by 
1) varying opinions and 2) non-aspect interference.
Accordingly, we design two types of contrastive objectives: 1) intra-aspect objective and 2) inter-aspect objective, 
where the former reinforces the differentiation between homogeneous and contrary opinions for an aspect, 
and the latter is account for distinguishing between the alteration for target/non-target aspects.

% $<$$X_{i,j}^{a,+},X_i^o$$>_j^+$
% $<$$X_{i,k}^{a,-},X_i^o$$>_k^-$

In intra-aspect contrastive learning, for each raw sample $X_i^o \in \mathbb{D}_o$ we construct positive pairs as $<$$X_{i,j}^{a,+},X_i^o$$>$, where $X_{i,j}^{a,+} \in \mathbb{D}_a$ is a pseudo instance in same polarity label with $X_i^o$, and negative pairs as $<$$X_{i,k}^{a,-},X_i^o$$>$, where $X_{i,k}^{a,-} \in \mathbb{D}_a$ comes from a different polarity label.
We encourage the system to learn nearer distance between positive pairs (\emph{Attract}), while enlarge the distance between negative pairs (\emph{Repel}):
\begin{equation}
\begin{aligned}
\label{cl_loss_intra}
\mathcal{L}_c^{ita} =& - \sum_j \log \frac{\exp[ \text{Sim}(\bm{r}(X_{i,j}^{a,+}),\bm{r}(X_i^o)) / \mu]}{\sum_k \exp[\text{Sim}(\bm{r}(X_{i,k}^{a,-}),\bm{r}(X_i^o)) / \mu]} \,, \\
&\text{Sim}(\bm{r}(a),\bm{r}(b))= \frac{\bm{r}(a)^T \bm{r}(b)}{||\bm{r}(a)|| \cdot ||\bm{r}(b)||} \,,
\end{aligned}
\end{equation}
where Sim($\cdot$) is the cosine similarity measurement, $\mu$ is a temperature factor.

Likewise, in inter-aspect contrastive learning, for each raw sample
we use the same positive pairs $<$$X_{i,j}^{a,+},X_i^o$$>$ as in $\mathcal{L}_c^{ita}$, representing inner-aspect changes, and construct negative pairs as $<$$X_{i,k}^{m},X_i^o$$>$, where $X_{i,k}^{m} \in \mathbb{D}_m$ representing outer-aspect changes:
\begin{equation}
\label{cl_loss_inter}
\mathcal{L}_c^{itr} = - \sum_j \log \frac{\exp[ \text{Sim}(\bm{r}(X_{i,j}^{a,+}),\bm{r}(X_i^o)) / \mu]}{\sum_k \exp[\text{Sim}(\bm{r}(X_{i,k}^{m}),\bm{r}(X_i^o)) / \mu]} \,.
\end{equation}
In Eq. (\ref{cl_loss_intra},\ref{cl_loss_inter}), $\bm{r}(X)$ is the feature representation $\bm{r}^f$ of the input $X$, which summarizes the opinion for the target aspect, i.e., opinion-guided one.
To further strengthen the learning effect, we propose a structure-guided contrastive method.
Technically, we instead use the pooled representation of syntax representation from USGCN, i.e., $\bm{r}^s=$Pool($\{\bm{r}^L_1,\cdots,\bm{r}^L_n\}$), which directly reflects the structural skeleton of the overall sentence.
Therefore we have in total four types of contrastive learning schemes: $\mathcal{L}_c^{ita\#o}$, $\mathcal{L}_c^{ita\#s}$, $\mathcal{L}_c^{itr\#o}$, $\mathcal{L}_c^{itr\#s}$, as illustrated in Figure \ref{CLs}.
We summarize the overall loss as:
\begin{equation}
\label{cl_loss}
% \begin{aligned}
\mathcal{L}_c(\mathbb{D}_o+\mathbb{D}_s) = \frac{1}{||\mathbb{D}_o+\mathbb{D}_s||} ( \lambda_{c1} \mathcal{L}_c^{ita\#o} + \lambda_{c2} \mathcal{L}_c^{ita\#s} 
 + \lambda_{c3} \mathcal{L}_c^{itr\#o} + \lambda_{c4} \mathcal{L}_c^{itr\#s} ) \,.
% \end{aligned}
\end{equation}

\paratitle{Jointly with Supervised Training.}
The unsupervised contrastive learning ($\mathcal{L}_c$ in Eq. \ref{cl_loss}) can join
1) the cross-entropy objective ($\mathcal{L}_e$ in Eq. \ref{cross-entropy}), 
or 2) the adversarial training objective ($\mathcal{L}_a$ in Eq. \ref{adv_loss}), i.e., $\mathcal{L}_{e+c}$ or $\mathcal{L}_{a+c}$.

\begin{table}[!t]
\caption{Statistics of datasets. 
}
% \vspace{-10pt}
\begin{center}
\resizebox{0.8\textwidth}{!}{
  \begin{tabular}{llcrrrr}
\toprule
 & Dataset & Domain	&Sentence	&Positive	&Neutral	&Negative \\
\midrule
\multirow{3}{*}{\bf Training set} 
& \multirow{2}{*}{SemEval}	&	Res	&	1,895	&	2,164	&	633		&805 \\
& 	&Lap		&1,365	&	987	&	460	&	866	\\
\cdashline{2-7}
& \normalsize{M}\scriptsize{A}\normalsize{M}\scriptsize{S}	&	Res	&	4,297	&	3,380	&	5,042	&	2,764	 \\
\hline
\multirow{3}{*}{\bf Developing set}
& \multirow{2}{*}{SemEval}	&	Res	&	84	&	70	&	54	&	26 \\
& 	&	Lap	&	98	&	57	&	27	&	66\\
\cdashline{2-7}
& \normalsize{M}\scriptsize{A}\normalsize{M}\scriptsize{S}	&	Res		&500	&	403		&604	&	325	\\
\hline
\multirow{5}{*}{\bf Testing set}
& \multirow{2}{*}{SemEval}	&	Res	&	600	&	728	&	196	&	196	 \\
& 	&	Lap	&	411	&	341	&	169	&	128 \\
\cdashline{2-7}
& \normalsize{M}\scriptsize{A}\normalsize{M}\scriptsize{S}	&	Res		&500	&	400		&607	&	329	 \\
\cdashline{2-7}
& \multirow{2}{*}{\normalsize{A}\scriptsize{RTS}}	&	Res		&492	&	1,953	&	473		&1,104	 \\
& 		&Lap	&	331	&	883	&	407	&	587	 \\
\hline
\end{tabular}
}
\end{center}
  \label{data-stat}
% \vspace{-10pt}
\end{table}

\section{Experiment}
\label{Experiment}

\subsection{Setups}

\subsubsection{Data and Resources}
Our experiments are based on the SemEval 2014 data \cite{pontiki-etal-2014-semeval}, which includes two subsets in two domains, \emph{Restaurant} and \emph{Laptop}.
Vanilla SemEval data evaluates in-house performances of ABSA models.
For the evaluation of robustness, we consider two datasets: {\normalsize{M}\scriptsize{A}\normalsize{M}\scriptsize{S}} \cite{jiang-etal-2019-challenge} and {\normalsize{A}\scriptsize{RTS}} \cite{xing-etal-2020-tasty}, which are described earlier at $\S$\ref{Related Work}.
Each dataset provides its own training/developing/testing sets.
Table \ref{data-stat} details the statistics of the datasets.
Note that here we cannot provide the data statistics of our constructed pseudo data ($\S$\ref{Synthetic Corpus Construction}), because of the dynamic process of of the data induction, i.e., we control the data quality by changing the thresholds, during which the data quantity is varying.
Besides, we obtain the syntax annotation\footnote{Following universal dependency v3.9.2.} of each sentence by a biaffine dependency parser \cite{BiaffineDozatM17}, which is trained on Penn Treebank corpus\footnote{\url{https://catalog.ldc.upenn.edu/LDC99T42}} and has an overall 93.4\% testing LAS.

\subsubsection{Comparing Methods.}
To make  a comprehensive comparisons with different neural network architectures, we consider a variety of types of existing ABSA systems as baselines\footnote{We run the experiments via their released codes.}.
% , including:

\begin{itemize}
    \item[$\blacktriangleright$] \textbf{LSTM-based model}. 1) \emph{TD-LSTM}. Tang et al. (2016) \cite{tang-etal-2016-effective} use two separate LSTMs to encode the forward and backward contexts of the target aspect (inclusive), and concatenate the last hidden states of the two LSTMs for making the sentiment classification.
    \item[$\blacktriangleright$] \textbf{Convolutional-based model}. 1) \emph{GCAE}. Xue et al. (2018) \cite{xue-li-2018-aspect} propose CNN-based model with gating mechanisms for selectively learning the sentiment features meanwhile keeping computational efficient with convolutions.
    \item[$\blacktriangleright$] \textbf{Attention-based models}. 
        1) \emph{MenNet}.  Tang et al. (2016)  \cite{tang-etal-2016-aspect} use a memory network to cache the sentential representations into external memory and then calculate the attention with the target aspect.
        2) \emph{AttLSTM}. Wang et al. (2016) \cite{wang-etal-2016-attention} quip the LSTM model with an attention mechanism, and concatenate the the aspect and word embeddings of each token for the final prediction.
        3) \emph{AOA}. Huang et al. (2018) \cite{HuangOC18} introduce an attention-over-attention network to jointly and explicitly capture the interaction between aspects and context sentences.
    \item[$\blacktriangleright$] \textbf{Capsule network}. 1) \emph{CapNet}. Jiang et al. (2019) \cite{jiang-etal-2019-challenge} employ a capsule network to encode the sentence as well as the aspect term so as to learn the encapsulated features of each sentiment polarity, and then take the routing algorithm to predict the polarity.
    \item[$\blacktriangleright$] \textbf{Syntax-based models}. 
        1) \emph{ASGCN}. Zhang et al. (2019) \cite{zhang-etal-2019-aspect} as the first effort utilize an aspect-specific GCN to encode the syntactic structure of the input sentence and then imposes an aspect-specific masking layer on its top to make prediction.
        2) \emph{TD-GAT}. Huang et al. (2019) \cite{huang-carley-2019-syntax} propose a multi-layer target-dependent graph attention network to explicitly encode the dependency tree information for better modeling the syntactic context of the target aspect.
        3) \emph{RGAT}. Wang et al. (2020) \cite{wang-etal-2020-relational} consider transform the original dependency tree into an aspect-oriented structure rooted at the target aspect, so as to prune the tree information for better sentiment prediction.
        4) \emph{RGCN}. Veyseh et al. (2020) \cite{pouran-ben-veyseh-etal-2020-improving} regulate the GCN-based representation vectors based on the dependency trees in order to benefit from the overall contextual importance scores of the words.
        
\end{itemize}

Also we explore the differences when additionally using the PLM representations, i.e., BERT\footnote{\url{https://github.com/google-research/bert}, uncased base version.}, including
PT+BERT \cite{xu-etal-2019-bert},
TD-LSTM+BERT,
CapNet+BERT,
ASGCN+BERT,
RGAT+BERT.

\subsubsection{Implementations and Evaluations.}
We use the pre-trained 300-d Glove embeddings \cite{PenningtonSM14}.
Transformer encoder has 768-d in 4 layers.
USGCN is with 300-d in 3 layers ($L$=3).
Syntax label embedding is with 100-d.
We use the mini-batch with a size of 16, training 10k iteration with early stopping.
We adopt the Adam optimizer with an initial learning rate as 1e-4, and a $\ell_2$ weight decay of 5e-5.
We apply 0.3 dropout ratio for word embeddings, and 0.1 for all other feature embeddings.
The thresholds $\theta_a, \theta_n$ and $\theta_m$ are set with 0.2, 0.25 and 0.85, respectively.
Based on preliminary experiments, $\lambda_a$=0.6, $\lambda_{c1}=\lambda_{c3}$=0.3 and $\lambda_{c2}=\lambda_{c4}$=0.2.
Following prior works, we use the accuracy to evaluate performance.
Each results of our model is from the average of ten times of running, and all the scores are presented statistically significant after paired T-test.
We fine-tune the hyper-parameters for all models on the validation set.
All experiments are conducted with a NVIDIA RTX GeForce 3090Ti GPU and 24 GB graphic memory.

\begin{table*}[!t]
  \caption{
  Testing results (accuracy) of ABSA systems on each test set, where the models are trained on raw SemEval data ($\mathbb{D}_o$) and the hybrid data with synthetic data (+$\mathbb{D}_s$), respectively.
In the brackets are the improvements by using additional synthetic training data.
$\dag$ means significance test with $p\le$0.05, and $\ddag$ means $p\le$0.03.
The underlined scores are the best results by using common cross-entropy training, and the bold scores are the best results using advanced training strategies.
}
% \vspace{-15pt}
\begin{center}
\resizebox{1.0\textwidth}{!}{
  \begin{tabular}{lllllllllll}
\toprule
\multirow{2}{*}{Test}& \multicolumn{4}{c}{SemEval}& \multicolumn{4}{c}{\normalsize{A}\scriptsize{RTS}}& \multicolumn{2}{c}{\normalsize{M}\scriptsize{A}\normalsize{M}\scriptsize{S}} \\
\cmidrule(r){2-5}\cmidrule(r){6-9}\cmidrule(r){10-11}
& \multicolumn{2}{c}{Restaurant}& \multicolumn{2}{c}{Laptop}& \multicolumn{2}{c}{Restaurant}& \multicolumn{2}{c}{Laptop}& \multicolumn{2}{c}{Restaurant} \\
\cmidrule(r){1-1}\cmidrule(r){2-3}\cmidrule(r){4-5}\cmidrule(r){6-7}\cmidrule(r){8-9}\cmidrule(r){10-11}
Train & \multicolumn{1}{c}{$\mathbb{D}_o$}&\multicolumn{1}{c}{+$\mathbb{D}_s$}& \multicolumn{1}{c}{$\mathbb{D}_o$}&\multicolumn{1}{c}{+$\mathbb{D}_s$}& \multicolumn{1}{c}{$\mathbb{D}_o$}&\multicolumn{1}{c}{+$\mathbb{D}_s$}& $\mathbb{D}_o$&\multicolumn{1}{c}{+$\mathbb{D}_s$}& \multicolumn{1}{c}{$\mathbb{D}_o$}&\multicolumn{1}{c}{+$\mathbb{D}_s$} \\	
% &R	&F1& P	&R	&F1& P	&R	&F1& P	&R	&F1& P	&R	&F1	\\
\midrule
\multicolumn{11}{l}{$\bullet$  \bf  w/o BERT}\\
MemNet &	75.18&	78.55(+3.37)$^\dag$&	64.42&	73.15(+8.73)$^\dag$&	33.34$^\dag$&	40.12(+6.78)$^\dag$&	32.34$^\dag$&	43.50(+11.16)$^\dag$&	39.85$^\dag$&	48.20(+8.35)$^\dag$ \\
AttLSTM &	75.98&	77.14(+1.16)$^\dag$&	67.55&	72.54(+4.99)$^\dag$&	26.52$^\dag$&	33.38(+6.86)$^\dag$&	31.87$^\dag$&	39.21(+7.34)$^\dag$&	30.21$^\dag$&	42.54(+12.33)$^\dag$ \\
TD-LSTM &	78.12&	78.92(+0.80)$^\dag$&	68.03&	73.68(+5.65)$^\dag$&	35.62$^\dag$&	43.85(+8.23)$^\dag$&	41.57$^\dag$&	52.52(+10.95)$^\dag$&	34.42$^\dag$&	44.92(+10.50)$^\dag$ \\
AOA &	79.32&	80.15(+0.83)$^\dag$&	72.60&	74.50(+1.90)$^\dag$&	30.02$^\dag$&	45.52(+15.50)$^\dag$&	40.35$^\dag$&	49.48(+9.13)$^\dag$&	32.36$^\dag$&	47.51(+15.15)$^\dag$ \\
GCAE &	79.53&	80.23(+0.70)$^\dag$&	73.15&	74.82(+1.67)$^\dag$&	36.58$^\dag$&	48.31(+11.73)$^\dag$&	35.66$^\dag$&	50.68(+15.02)$^\dag$&	40.25$^\dag$&	50.89(+10.64)$^\dag$ \\
CapNet &	80.16&	80.58(+0.42)$^\dag$&	73.54&	75.21(+1.67)$^\dag$&	38.89$^\dag$&	44.65(+5.76)$^\dag$&	45.32$^\dag$&	54.51(+9.19)$^\dag$&	38.16$^\dag$&	50.52(+12.36)$^\dag$ \\
\hdashline
ASGCN &	80.86&	81.39(+0.53)$^\dag$&	74.61&	75.98(+1.37)$^\dag$&	44.20$^\dag$&	52.47(+8.27)$^\dag$&	59.24$^\dag$&	66.77(+7.53)$^\dag$&	45.25$^\dag$&	52.02(+6.77)$^\dag$ \\
TD-GAT &	81.20&	82.07(+0.87)$^\dag$&	74.00&	75.34(+1.34)$^\dag$&	40.32$^\dag$&	49.15(+8.83)$^\dag$&	53.38$^\dag$&	60.85(+7.47)$^\dag$&	43.10$^\dag$&	52.77(+9.67)$^\dag$ \\
RGAT &	82.12&	82.65(+0.53)$^\dag$&	75.20&	75.72(+0.52)$^\dag$&	41.73$^\dag$&	51.58(+9.85)$^\dag$&	54.91$^\dag$&	62.34(+7.43)$^\dag$&	41.89$^\dag$&	51.50(+9.61)$^\dag$ \\
\hdashline
Ours($\mathcal{L}_{e}$)& 	\underline{82.85}$^\ddag$& 	\underline{83.13}(+0.28)$^\ddag$& 	\underline{76.22}$^\ddag$& 	\underline{76.85}(+0.63)$^\ddag$& 	\underline{46.57}$^\ddag$& 	\underline{55.58}(+9.01)$^\ddag$& 	\underline{61.33}$^\ddag$& 	\underline{69.12}(+7.79)$^\ddag$& 	\underline{47.25}$^\ddag$& 	\underline{55.34}(+8.09)$^\ddag$ \\
Ours($\mathcal{L}_{a}$)&	- &	83.52$^\ddag$&	-&	77.12$^\ddag$&	-&	58.61$^\ddag$&	-&	70.53$^\ddag$&	-&	56.12$^\ddag$ \\
Ours($\mathcal{L}_{e+c}$)&	-&	83.98$^\ddag$&	-&	77.07$^\ddag$&	-&	58.20$^\ddag$&	-&	70.68$^\ddag$&	-&	56.53$^\ddag$ \\
Ours($\mathcal{L}_{a+c}$)&	-&	\textbf{84.45}$^\ddag$&	-&	\textbf{77.53}$^\ddag$&	-&	\textbf{60.39}$^\ddag$&	-&	\textbf{71.21}$^\ddag$&	-&	\textbf{57.02}$^\ddag$ \\
\hdashline
\textbf{\em Avg.} &	79.53&	80.48(+0.95)&	71.93&	74.78(+2.85)&	37.38&	46.46(+9.08)&	45.60&	54.90(+9.30)&	39.27&	49.62(+10.35) \\
\hline
\multicolumn{11}{l}{$\bullet$  \bf  w/ BERT}\\
BERT &	83.04$^\dag$&	84.66(+1.62)$^\dag$&	77.59$^\dag$&	78.69(+1.10)$^\dag$&	66.23$^\dag$&	75.35(+9.12)$^\dag$&	62.42$^\dag$&	69.55(+7.13)$^\dag$&	51.32$^\dag$&	56.85(+5.53)$^\dag$ \\
TD-LSTM+BERT&	84.51$^\dag$&	85.28(+0.77)$^\dag$&	77.98$^\dag$&	78.86(+0.88)$^\dag$&	68.45$^\dag$&	75.56(+7.11)$^\dag$&	63.26$^\dag$&	69.63(+6.37)$^\dag$&	50.67$^\dag$&	57.12(+6.45)$^\dag$ \\
CapNet+BERT &	85.48$^\dag$&	86.04(+0.56)$^\dag$&	77.12$^\dag$&	79.30(+2.18)$^\dag$&	69.36$^\dag$&	77.48(+8.12)$^\dag$&	64.01$^\dag$&	70.21(+6.20)$^\dag$&	52.23$^\dag$&	57.14(+4.91)$^\dag$ \\
PT+BERT  &	86.40$^\dag$&	86.75(+0.35)$^\dag$&	78.06$^\dag$&	79.12(+1.06)$^\dag$&	71.41$^\dag$&	77.59(+6.18)$^\dag$&	65.23$^\dag$&	72.02(+6.79)$^\dag$&	54.16$^\dag$&	58.69(+4.53)$^\dag$ \\
\hdashline
ASGCN+BERT&	86.82$^\dag$&	87.24(+0.42)$^\dag$&	78.53$^\dag$&	79.53(+1.00)$^\dag$&	73.48$^\dag$&	78.18(+4.70)$^\dag$&	67.63$^\dag$&	72.85(+5.22)$^\dag$&	55.42$^\dag$&	59.48(+4.06)$^\dag$ \\
RGAT+BERT 2020&	86.60$^\dag$&	87.03(+0.43)$^\dag$&	78.20$^\dag$&	79.38(+1.18)$^\dag$&	72.83$^\dag$&	78.25(+5.42)$^\dag$&	67.28$^\dag$&	71.35(+4.07)$^\dag$&	55.84$^\dag$&	60.52(+4.68)$^\dag$ \\
\hdashline
Ours+BERT($\mathcal{L}_{e}$)& 	\underline{87.05}$^\ddag$& 	\underline{87.15}(+0.10)$^\ddag$& 	\underline{79.61}$^\ddag$& 	\underline{80.28}(+0.67)$^\ddag$& 	\underline{75.01}$^\ddag$& 	\underline{80.65}(+5.64)$^\ddag$& 	\underline{68.78}$^\ddag$& 	\underline{73.89}(+5.11)$^\ddag$& 	\underline{57.03}$^\ddag$& 	\underline{62.37}(+5.34)$^\ddag$ \\
Ours+BERT($\mathcal{L}_{a}$)&	-&	87.53$^\ddag$&	-&	80.85$^\ddag$&	-&	81.95$^\ddag$&	-&	74.52$^\ddag$&	-&	63.07$^\ddag$ \\
Ours+BERT($\mathcal{L}_{e+c}$)&	-&	87.49$^\ddag$&	-&	80.34$^\ddag$&	-&	81.42$^\ddag$&	-&	74.36$^\ddag$&	-&	63.24$^\ddag$ \\
Ours+BERT($\mathcal{L}_{a+c}$)&	-&	\textbf{87.87}$^\ddag$&	-&	\textbf{81.26}$^\ddag$&	-&	\textbf{82.38}$^\ddag$&	-&	\textbf{75.65}$^\ddag$&	-&	\textbf{63.58}$^\ddag$ \\
\hdashline
\textbf{\em Avg.} &	85.70&	86.31(+0.61)&	78.16&	79.31(+1.15)&	70.97&	77.58(+6.61)&	65.52&	71.36(+5.84)&	53.81&	58.88(+5.07) \\
\hline
\end{tabular}
}
\end{center}
  \label{result-main1}
% \vspace{-10pt}
\end{table*}

\begin{table}[!t]
\caption{
Training with advanced strategies.
`w/o s.l.' and `w/o a.': removing syntax label ($\bm{x}^e_{i,j}$) and aspect embedding ($\bm{r}^{asp}$) from USGCN (Eq. \ref{USGCN-attention}), respectively.
`w/o Trm': replacing Transformer encoder with BiLSTM.
}
% \vspace{-10pt}
\begin{center}
\resizebox{0.6\columnwidth}{!}{
  \begin{tabular}{llllll}
\toprule
 & \multicolumn{2}{c}{SemEval}& \multicolumn{2}{c}{\normalsize{A}\scriptsize{RTS}}& \multicolumn{1}{c}{\normalsize{M}\scriptsize{A}\normalsize{M}\scriptsize{S}} \\
\cmidrule(r){2-3}\cmidrule(r){4-5}\cmidrule(r){6-6}
& \multicolumn{1}{c}{Res}& \multicolumn{1}{c}{Lap}& \multicolumn{1}{c}{Res}& \multicolumn{1}{c}{Lap}& \multicolumn{1}{c}{Res} \\
\midrule
\multicolumn{6}{l}{$\bullet$  \bf Training with $\mathcal{L}_{e+c}$} \\
TD-LSTM	&79.65$^\dag$	&74.88$^\dag$	&46.85$^\dag$	&54.32$^\dag$	&48.55$^\dag$ \\
GCAE	&81.42$^\dag$	&75.41$^\dag$	&50.47$^\dag$	&52.40$^\dag$	&52.02$^\dag$ \\
CapNet	&81.69$^\dag$	&76.10$^\dag$	&47.78$^\dag$	&56.85$^\dag$	&52.63$^\dag$ \\
ASGCN	&82.02$^\dag$	&76.49$^\dag$	&55.46$^\dag$	&67.85$^\dag$	&54.22$^\dag$ \\
RGAT	&83.02$^\dag$	&76.28$^\dag$	&53.91$^\dag$	&63.47$^\dag$	&53.28$^\dag$ \\
\hdashline
Ours	&\bf83.98$^\ddag$	&\bf77.07$^\ddag$	&\bf58.20$^\ddag$	&\bf70.68$^\ddag$	&\bf56.53$^\ddag$ \\
Ours(w/o s.l.)	&82.16$^\ddag$	&76.50$^\ddag$	&54.56$^\ddag$	&69.56$^\ddag$	&54.70$^\ddag$ \\
Ours(w/o a.)	&83.65$^\ddag$	&76.95$^\ddag$	&57.92$^\ddag$	&70.44$^\ddag$	&55.00$^\ddag$ \\
Ours(w/o Trm)	&83.31$^\ddag$	&76.88$^\ddag$	&57.70$^\ddag$	&70.32$^\ddag$	&55.13$^\ddag$ \\
\hdashline
\textbf{\em Avg.}	&82.32	&76.28	&53.65	&63.99	&53.56 \\
\hline
\multicolumn{6}{l}{$\bullet$  \bf Training with $\mathcal{L}_{a+c}$} \\
TD-LSTM	&80.67$^\dag$	&75.23$^\dag$	&48.61$^\dag$	&55.11$^\dag$	&49.34$^\dag$ \\
GCAE	&81.83$^\dag$	&75.83$^\dag$	&52.42$^\dag$	&53.22$^\dag$	&53.83$^\dag$ \\
CapNet	&81.96$^\dag$	&76.76$^\dag$	&50.03$^\dag$	&58.23$^\dag$	&52.95$^\dag$ \\
ASGCN	&82.35$^\dag$	&76.89$^\dag$	&56.74$^\dag$	&68.50$^\dag$	&54.87$^\dag$ \\
RGAT	&83.64$^\dag$	&76.79$^\dag$	&55.62$^\dag$	&64.72$^\dag$	&53.94$^\dag$ \\
\hdashline
Ours	&\bf 84.45$^\ddag$	&\bf77.53$^\ddag$	&\bf60.39$^\ddag$	&\bf71.21$^\ddag$	&\bf57.02$^\ddag$ \\
Ours(w/o s.l.)	&82.58$^\ddag$	&76.95$^\ddag$	&56.82$^\ddag$	&70.14$^\ddag$	&55.06$^\ddag$ \\
Ours(w/o a.)	&83.90$^\ddag$	&77.32$^\ddag$	&59.35$^\ddag$	&71.02$^\ddag$	&56.79$^\ddag$ \\
Ours(w/o Trm)	&83.72$^\ddag$	&77.06$^\ddag$	&58.45$^\ddag$	&70.75$^\ddag$	&56.21$^\ddag$ \\
\hdashline
\textbf{\em Avg.}	&82.79	&76.71	&55.38	&64.77	&54.45 \\
\hline
\end{tabular}
}
\end{center}
  \label{blend-train}
% \vspace{-10pt}
\end{table}

\subsection{Main Results}

We consider two types of evaluations, i.e., training based on SemEval and {\normalsize{M}\scriptsize{A}\normalsize{M}\scriptsize{S}} data, where the former is with challenge-less data, and the latter is with challenge-aware data.
Based on both two setups, we evaluate the effectiveness of our model, corpus and training strategies by making comparisons with baselines.
In first setup, ABSA models are trained on SemEval data and evaluated on different testing sets (i.e., SemEval, {\normalsize{A}\scriptsize{RTS}} and {\normalsize{M}\scriptsize{A}\normalsize{M}\scriptsize{S}}).
In second setup, we train ABSA models on {\normalsize{M}\scriptsize{A}\normalsize{M}\scriptsize{S}} and then perform testing.

\begin{table*}[!t]
  \caption{
  Fine-grained robustness testing performances on each subset of {\normalsize{A}\scriptsize{RTS}} data.
}
% \vspace{-10pt}
\begin{center}
\resizebox{1.0\textwidth}{!}{
  \begin{tabular}{lllllllll}
\toprule
Test& \multicolumn{2}{c}{\normalsize{R}\scriptsize{EV}\normalsize{T}\scriptsize{GT}}& \multicolumn{2}{c}{\normalsize{R}\scriptsize{EV}\normalsize{N}\scriptsize{ON}}& \multicolumn{2}{c}{\normalsize{A}\scriptsize{DD}\normalsize{D}\scriptsize{IFF}}& \multicolumn{2}{c}{\normalsize{R}\scriptsize{WT}\normalsize{B}\scriptsize{G}} \\
\cmidrule(r){1-1}\cmidrule(r){2-3}\cmidrule(r){4-5}\cmidrule(r){6-7}\cmidrule(r){8-9}
Train& \multicolumn{1}{c}{$\mathbb{D}_o$}&\multicolumn{1}{c}{+$\mathbb{D}_s$}& \multicolumn{1}{c}{$\mathbb{D}_o$}&\multicolumn{1}{c}{+$\mathbb{D}_s$}& \multicolumn{1}{c}{$\mathbb{D}_o$}&\multicolumn{1}{c}{+$\mathbb{D}_s$}& $\mathbb{D}_o$&\multicolumn{1}{c}{+$\mathbb{D}_s$} \\	
\midrule
\multicolumn{9}{l}{$\bullet$  \bf  w/o BERT }\\
MemNet	&27.54$^\dag$	&80.73(+53.19)$^\dag$	&73.65$^\dag$	&84.46(+10.81)$^\dag$	&60.71$^\dag$	&75.18(+14.47)$^\dag$	&77.50$^\dag$	&80.33(+2.83)$^\dag$ \\
AttLSTM	&28.98$^\dag$	&82.98(+54.00)$^\dag$	&61.26$^\dag$	&77.26(+16.00)$^\dag$	&52.32$^\dag$	&75.98(+23.66)$^\dag$	&69.64$^\dag$	&84.44(+14.80)$^\dag$ \\
AOA	&30.51$^\dag$	&84.36(+53.85)$^\dag$	&73.95$^\dag$	&84.13(+10.18)$^\dag$	&63.51$^\dag$	&72.55(+9.04)$^\dag$	&70.54$^\dag$	&82.36(+11.82)$^\dag$ \\
GCAE	&33.02$^\dag$	&85.15(+52.13)$^\dag$	&75.02$^\dag$	&85.63(+10.61)$^\dag$	&63.72$^\dag$	&76.45(+12.73)$^\dag$	&74.27$^\dag$	&84.67(+10.40)$^\dag$ \\
CapNet	&30.15$^\dag$	&85.37(+55.22)$^\dag$	&76.36$^\dag$	&84.69(+8.33)$^\dag$	&57.65$^\dag$	&75.59(+17.94)$^\dag$	&78.56$^\dag$	&86.85(+8.29)$^\dag$ \\
ASGCN	&34.78$^\dag$	&86.76(+51.98)$^\dag$	&79.50$^\dag$	&88.51(+9.01)$^\dag$	&70.88$^\dag$	&78.86(+7.98)$^\dag$	&80.63$^\dag$	&90.04(+9.41)$^\dag$ \\
RGAT	&37.05$^\dag$	&87.26(+50.21)$^\dag$	&81.15$^\dag$	&87.03(+5.88)$^\dag$	&67.05$^\dag$	&79.48(+12.43)$^\dag$	&78.15$^\dag$	&89.85(+9.70)$^\dag$ \\
\hdashline
Ours($\mathcal{L}_{e}$)	& \underline{40.41}$^\ddag$	& \underline{88.33}(+47.92)$^\ddag$	& \underline{80.62}$^\ddag$	& \underline{90.52}(+9.90)$^\ddag$	& \underline{74.66}$^\ddag$	& \underline{81.56}(+6.90)$^\ddag$	& \underline{82.84}$^\ddag$	& \underline{92.54}(+9.70)$^\ddag$ \\
Ours($\mathcal{L}_{a}$)	&-	&89.51$^\ddag$	&-	&91.30$^\ddag$	&-	&82.69$^\ddag$	&-	&92.98$^\ddag$ \\
Ours($\mathcal{L}_{e+c}$)	&-	&89.28$^\ddag$	&-	&90.89$^\ddag$	&-	&81.98$^\ddag$	&-	&92.71$^\ddag$ \\
Ours($\mathcal{L}_{a+c}$)	&-	&\textbf{90.42}$^\ddag$	&-	&\textbf{91.65}$^\ddag$	&-	&\textbf{83.13}$^\ddag$	&-	&\textbf{93.45}$^\ddag$ \\
\hdashline
\textbf{\em Avg.}	&32.81	&85.12(+52.31)	&75.44	&85.53(+10.09)	&63.81	&76.96(+13.15)	&77.02	&86.63(+9.61) \\
\hline
\multicolumn{9}{l}{$\bullet$  \bf  w/ BERT }\\
BERT	&63.00$^\dag$	&84.15(+21.15)$^\dag$	&83.33$^\dag$	&86.33(+3.00)$^\dag$	&79.20$^\dag$	&85.79(+6.59)$^\dag$	&81.36$^\dag$	&82.20(+0.84)$^\dag$ \\
TD-LSTM+BERT	&67.32$^\dag$	&85.85(+18.53)$^\dag$	&80.68$^\dag$	&88.15(+7.47)$^\dag$	&79.35$^\dag$	&86.22(+6.87)$^\dag$	&80.30$^\dag$	&88.41(+8.11)$^\dag$ \\
CapNet+BERT	&71.87$^\dag$	&87.74(+15.87)$^\dag$	&78.55$^\dag$	&86.48(+7.93)$^\dag$	&77.86$^\dag$	&85.96(+8.10)$^\dag$	&83.02$^\dag$	&87.05(+4.03)$^\dag$ \\
PT+BERT	&72.83$^\dag$	&84.33(+11.50)$^\dag$	&81.76$^\dag$	&88.87(+7.11)$^\dag$	&80.27$^\dag$	&87.77(+7.50)$^\dag$	&82.48$^\dag$	&84.68(+2.20)$^\dag$ \\
ASGCN+BERT	&74.51$^\dag$	&89.76(+15.25)$^\dag$	&85.12$^\dag$	&90.35(+5.23)$^\dag$	&82.52$^\dag$	&88.31(+5.79)$^\dag$	&83.85$^\dag$	&91.68(+7.83)$^\dag$ \\
RGAT+BERT	&75.68$^\dag$	&90.48(+14.80)$^\dag$	&83.38$^\dag$	&91.21(+7.83)$^\dag$	&80.45$^\dag$	&87.88(+7.43)$^\dag$	&84.64$^\dag$	&92.45(+7.81)$^\dag$ \\
\hdashline
Ours+BERT($\mathcal{L}_{e}$)	& \underline{78.02}$^\ddag$	& \underline{91.32}(+13.30)$^\ddag$	& \underline{86.32}$^\ddag$	& \underline{92.86}(+6.54)$^\ddag$	& \underline{82.14}$^\ddag$	& \underline{89.68}(+7.54)$^\ddag$	& \underline{85.45}$^\ddag$	& \underline{93.52}(+8.07)$^\ddag$ \\
Ours+BERT($\mathcal{L}_{a}$)	&-	&92.45$^\ddag$	&-	&93.45$^\ddag$	&-	&90.46$^\ddag$	&-	&94.22$^\ddag$ \\
Ours+BERT($\mathcal{L}_{e+c}$)	&-	&92.04$^\ddag$	&-	&93.11$^\ddag$	&-	&90.35$^\ddag$	&-	&94.06$^\ddag$ \\
Ours+BERT($\mathcal{L}_{a+c}$)	&-	&\textbf{93.12}$^\ddag$	&-	&\textbf{93.76}$^\ddag$	&-	&\textbf{90.85}$^\ddag$	&-	&\textbf{95.18}$^\ddag$ \\
\hdashline
\textbf{\em Avg.}	&72.31	&87.89(+15.58)	&83.00	&89.41(+6.41)	&80.27	&87.41(+7.14)	&83.14	&89.05(+5.91) \\
\hline
\end{tabular}
}
\end{center}
  \label{result-ARTS}
% \vspace{-10pt}
\end{table*}

\subsubsection{Training Based on SemEval data.}
In Table \ref{result-main1} we show the main performances on each test set.
Besides of the SemEval training data (denoted as $\mathbb{D}_o$), we also consider the training with the additional synthetic data (i.e., $\mathbb{D}_o+\mathbb{D}_s$).
We correspondingly gain multiple observations.
The starting observation is that all the ABSA models (even the state-of-the-art ones) trained on SemEval data can drop significantly when testing on the challenging data ({\normalsize{A}\scriptsize{RTS}} and {\normalsize{M}\scriptsize{A}\normalsize{M}\scriptsize{S}}).
This reveals the imperative to enhance the ABSA robustness.

The second observation is about the ABSA models.
We see that baselines with different kinds of neural architectures show different generalization capabilities.
For example, the syntax-aware models not only give stronger performances on in-house test than other types, but also consistently preserve better robustness.
This confirms the prior findings in \cite{xing-etal-2020-tasty} that explicitly modeling the aspect-position information (such as syntax-aware model) leads to superior robustness.
More significantly, our proposed syntax-aware neural system shows the best performances in both in-house and out-of-house test, i.e., stronger generalization ability.
At the same time we find that the attention-based models actually give quite lower robustness performances, while using the pre-trained BERT representations the drops on the challenging testing data by ABSA models relieves greatly, i.e., PLM can help to enhance the ABSA robustness.

Furthermore, training additionally with the synthetic data all the ABSA models obtain improved performances than the counterparts (marked in the brackets) in both in-house and out-of-house test across all the testing sets universally.
Especially we see that the robustness performances on out-of-house test data are substantially enhanced.
These boosts are more obvious when the BERT PLM is not used.
This reveals the significance to enrich the training data with more additional challenging signals for robust ABSA.
Also this again proves the helps of PLM for improving ABSA robustness \cite{jiang-etal-2019-challenge,xing-etal-2020-tasty}.

Last but not least, we see that using the training paradigm based on our pseudo corpus, our system can receive further enhancements consistently on all the test sets.
Specifically, we consider different combination of training mechanisms, i.e., $\mathcal{L}_e$, $\mathcal{L}_a$, $\mathcal{L}_{e+c}$ and $\mathcal{L}_{e+a}$.
It shows that both adversarial training $\mathcal{L}_a$ and contrastive learning $\mathcal{L}_{e+c}$ can result in better performances than the basic cross-entropy training $\mathcal{L}_e$, while integrating both two training strategies ($\mathcal{L}_{e+a}$) our model gives the best effects.
Also, we see from Table \ref{blend-train} that all comparing baselines achieve consistent improvements on robustness when the advanced training strategies ($\mathcal{L}_{e+c}$ and $\mathcal{L}_{e+a}$) are equipped with pseudo data.

Table \ref{blend-train} shows the ablation results of our proposed models.
Removing the aspect from the unified modeling with syntax in USGCN shows inferior accuracies.
Without encoding the dependency syntax knowledge, our USGCN encoder results in significant performance drops, which reflects the importance to model the universal syntax for ABSA.
Further, without using Transformer encoder, we also witness the downgraded performances.
But each of our ablated model still outperforms the best baseline, i.e., ASGCN model only encodes the dependency edge information.

\begin{table}[!t]
  \caption{
  Robustness test results where models are trained on {\normalsize{M}\scriptsize{A}\normalsize{M}\scriptsize{S}} (denoted as $\mathbb{D}_o$) and with additional the pseudo data (+$\mathbb{D}_s$).
}
% \vspace{-12pt}
\begin{center}
\resizebox{0.64\textwidth}{!}{
  \begin{tabular}{lllll}
\toprule
Test& \multicolumn{2}{c}{\normalsize{M}\scriptsize{A}\normalsize{M}\scriptsize{S}} &\multicolumn{2}{c}{\normalsize{A}\scriptsize{RTS}} \\
\cmidrule(r){1-1}\cmidrule(r){2-3}\cmidrule(r){4-5}
Train& \multicolumn{1}{c}{$\mathbb{D}_o$}&\multicolumn{1}{c}{+$\mathbb{D}_s$}& \multicolumn{1}{c}{$\mathbb{D}_o$}&\multicolumn{1}{c}{+$\mathbb{D}_s$}\\	
\midrule
\multicolumn{5}{l}{$\bullet$  \bf  w/o BERT }\\
MemNet	&73.24$^\dag$	&75.85(+2.61)$^\dag$	&69.67$^\dag$	&74.15(+4.48)$^\dag$ \\
AttLSTM	&70.53$^\dag$	&74.12(+3.59)$^\dag$	&65.25$^\dag$	&70.45(+5.20)$^\dag$ \\
TD-LSTM	&74.59$^\dag$	&76.27(+1.68)$^\dag$	&69.51$^\dag$	&72.36(+2.85)$^\dag$ \\
AOA	&75.27$^\dag$	&77.54(+2.27)$^\dag$	&68.33$^\dag$	&71.85(+3.52)$^\dag$ \\
GCAE	&75.82$^\dag$	&77.80(+1.98)$^\dag$	&71.52$^\dag$	&76.44(+4.92)$^\dag$ \\
CapNet	&75.77$^\dag$	&77.36(+1.59)$^\dag$	&73.78$^\dag$	&77.38(+3.60)$^\dag$ \\
ASGCN	&76.95$^\dag$	&79.45(+2.50)$^\dag$	&75.12$^\dag$	&78.57(+3.45)$^\dag$ \\
TD-GAT	&78.54$^\dag$	&80.06(+1.52)$^\dag$	&75.69$^\dag$	&78.02(+2.33)$^\dag$ \\
RGAT	&79.09$^\dag$	&81.20(+2.11)$^\dag$	&76.24$^\dag$	&79.24(+3.00)$^\dag$ \\
\hdashline
Ours($\mathcal{L}_{e}$)	&\underline{80.65}$^\ddag$	&\underline{82.63}(+1.98)$^\ddag$	&\underline{77.50}$^\ddag$	&\underline{80.48}(+2.98)$^\ddag$ \\
Ours($\mathcal{L}_{a}$)	&-	&83.48$^\ddag$	&-	&82.02$^\ddag$ \\
Ours($\mathcal{L}_{e+c}$)	&-	&82.92$^\ddag$	&-	&81.25$^\ddag$ \\
Ours($\mathcal{L}_{a+c}$)	&-	&\bf84.17$^\ddag$	&-	&\bf82.44$^\ddag$ \\
\hdashline
\textbf{\em Avg.}	&76.04	&78.23(+2.19)	&72.26	&75.89(+3.63) \\
\hline
\multicolumn{5}{l}{$\bullet$  \bf  w/ BERT }\\
CapNet+BERT	&83.39$^\dag$	&84.72(+1.33)$^\dag$	&79.18$^\dag$	&82.48(+3.30)$^\dag$ \\
BERT+Xu	&82.52$^\dag$	&84.65(+2.13)$^\dag$	&79.38$^\dag$	&82.67(+3.29)$^\dag$ \\
PT+BERT	&83.10$^\dag$	&84.88(+1.78)$^\dag$	&80.07$^\dag$	&83.24(+3.17)$^\dag$ \\
RGAT+BERT	&83.93$^\dag$	&85.15(+1.22)$^\dag$	&80.48$^\dag$	&83.45(+2.97)$^\dag$ \\
\hdashline
Ours+BERT($\mathcal{L}_{e}$)	&\underline{84.23}$^\ddag$	&\underline{86.04}(+1.81)$^\ddag$	&\underline{81.56}$^\ddag$	&\underline{84.66}(+3.10)$^\ddag$ \\
Ours+BERT($\mathcal{L}_{a}$)	&-	&86.78$^\ddag$	&-	&85.47$^\ddag$ \\
Ours+BERT($\mathcal{L}_{e+c}$)	&-	&86.45$^\ddag$	&-	&85.02$^\ddag$ \\
Ours+BERT($\mathcal{L}_{a+c}$)	&-	&\bf87.12$^\ddag$	&-	&\bf86.93$^\ddag$ \\
\hdashline
\textbf{\em Avg.}	&83.43	&85.09(+1.66)	&80.13	&83.30(+3.17) \\
\hline
\end{tabular}
}
\end{center}
  \label{result-MAMS}
% \vspace{-10pt}
\end{table}

\subsubsection{Fine-grained Robustness Testing.}
In the {\normalsize{A}\scriptsize{RTS}} challenging test set there are three subsets ({\normalsize{R}\scriptsize{EV}\normalsize{T}\scriptsize{GT}}, {\normalsize{R}\scriptsize{EV}\normalsize{N}\scriptsize{ON}} and {\normalsize{A}\scriptsize{DD}\normalsize{D}\scriptsize{IFF}}), each of which evaluates the ABSA robustness from different aspects.
For example,
{\normalsize{R}\scriptsize{EV}\normalsize{T}\scriptsize{GT}} measures if a model can correctly bind target aspect to its critical opinion clues,
{\normalsize{R}\scriptsize{EV}\normalsize{N}\scriptsize{ON}} detects the sensitivity of a model to the sentiment change of non-target aspects,
and {\normalsize{A}\scriptsize{DD}\normalsize{D}\scriptsize{IFF}} testifies if a model is robust to the existence of non-target aspect.
In Table \ref{result-ARTS} we show the specific performances w.r.t each of the {\normalsize{A}\scriptsize{RTS}} subset (\emph{Restaurant}).
To further evaluate the robustness to the change of trivial background contexts, we additionally build a test set\footnote{We first derive the pseudo data as in $\S$\ref{Background Rewriting}, and then manually inspect the data to ensure the quality.} {\normalsize{R}\scriptsize{WT}\normalsize{B}\scriptsize{G}}.

From the results in Table \ref{result-ARTS} we learn that almost all ABSA models give most significant accuracy drops on {\normalsize{R}\scriptsize{EV}\normalsize{T}\scriptsize{GT}} than on other subsets, which we regard as the major bottleneck of robust ABSA.
However, our pseudo training data helps to substantially compensate such drops on {\normalsize{R}\scriptsize{EV}\normalsize{T}\scriptsize{GT}} for all these models, e.g., an average of 52.31\% accuracy increase.
For other robust testing subset, our synthetic data also provides helps, e.g., around 10\% accuracy increase.
Likewise, our proposed ABSA model always shows better results than baselines.
Interestingly, with BERT PLM information, the drops on robustness test of each model are much relieved, and correspondingly the positive effects from our pseudo data are not that prominent.
But we still see that introducing of two advanced training strategies in our model steadily leads to further improvements.

\subsubsection{Training Based on MAMS data.}
Table \ref{result-MAMS} shows the performances of ABSA models trained on {\normalsize{M}\scriptsize{A}\normalsize{M}\scriptsize{S}} data.
We see that the earlier viewpoint is verified that training with more challenging training data the robustness can be greatly improved, i.e., the gaps of the accuracy between in-house testing (on {\normalsize{M}\scriptsize{A}\normalsize{M}\scriptsize{S}}) and out-of-house testing (on {\normalsize{A}\scriptsize{RTS}}) are not as significant as those observed in Table \ref{result-main1}.
This conclusion is further supported by the observation that additionally using our pseudo data ($\mathbb{D}_o+\mathbb{D}_s$) helps to obtain much limited improvements.
The rest of the observations are kept same with that in Table \ref{result-main1}, i.e., 1) syntax-aware models show stronger capabilities, while 2) our proposed model gives the best performances, and 3) PLM helps achieve better robustness.

\section{Analysis and Discussion}
\label{Analysis and Discussion}

In prior experiments, we show the effectiveness of the proposed ABSA model, the synthetic training corpus and the advanced training paradigms for better ABSA robustness.
In this section, we take further steps, exploring the factors influencing the performances on these three aspects.

\begin{figure}[!t]
\centering
\includegraphics[width=0.6\columnwidth]{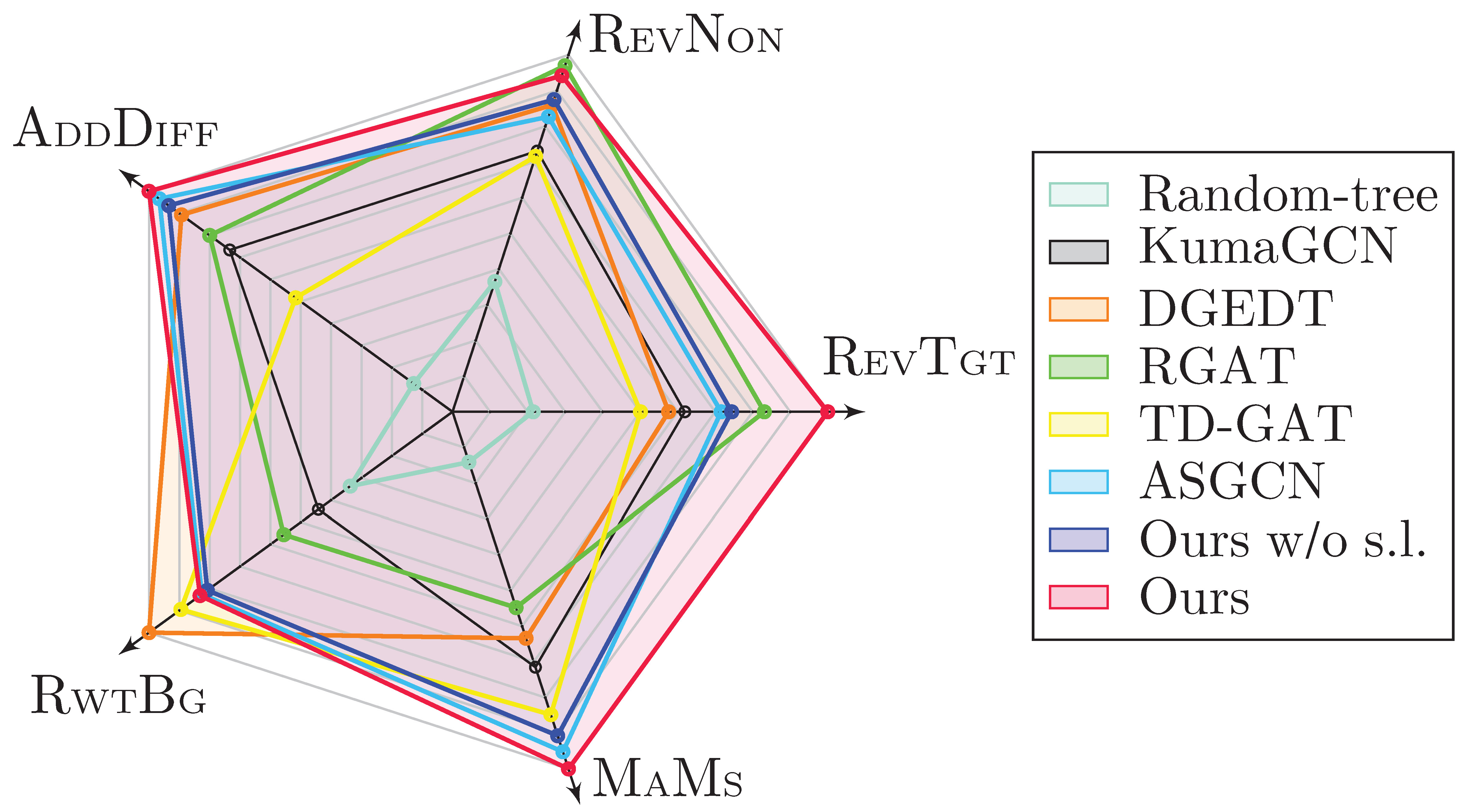}
\caption{
Radar map of the performances by different syntax-aware ABSA models on each specific robust test.
}
\label{radar}
\end{figure}

\begin{figure}[!t]
\centering
\includegraphics[width=0.6\columnwidth]{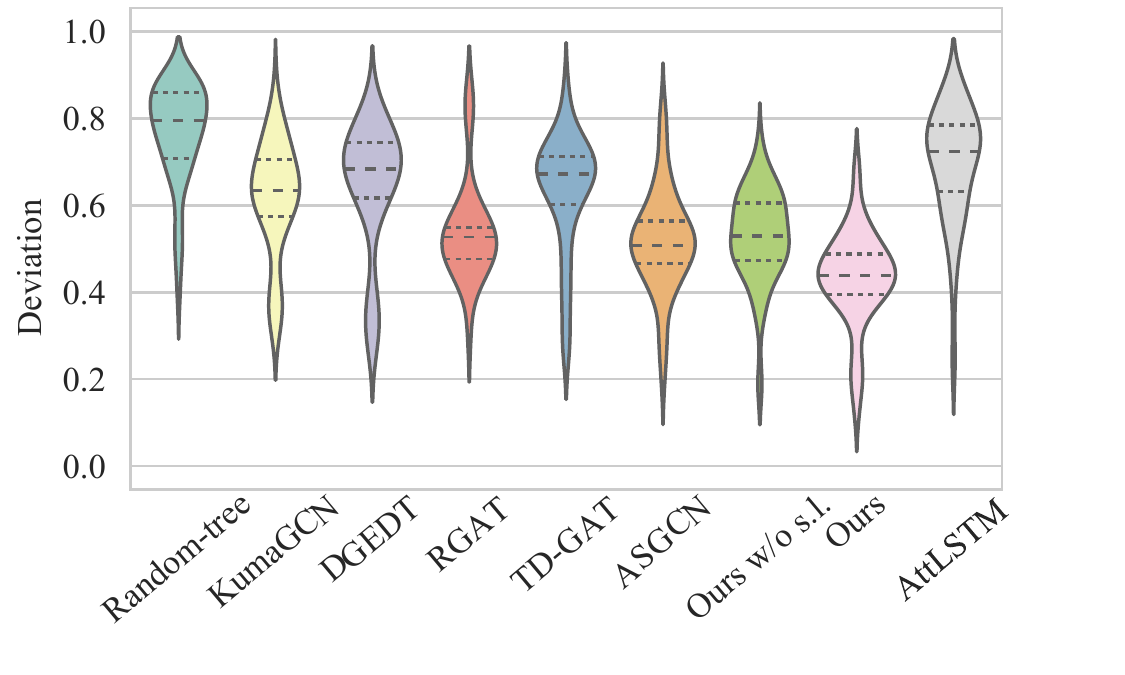}
\caption{
Model deviations on the faithfulness of aspect's opinion clues.
}
\label{violinplot}
\end{figure}

\subsection{Model Evaluation}

Above we show that the syntax integration and PLM greatly enhance the robustness.
Here we try to find answers for the following questions.

% \vspace{-7pt}
\begin{itemize}
    \item[\textbf{Q1}:] \emph{How much will the robustness score vary across the different syntax integration methods?}
    \item[\textbf{Q2}:] \emph{Why does syntax-based model improve robustness?}
    \item[\textbf{Q3}:] \emph{To what extent does syntax quality influence robustness?}
    \item[\textbf{Q4}:] \emph{Can Stronger PLM bring better ABSA robustness?}
\end{itemize}
% \vspace{-8pt}

\subsubsection{Performances of Different Syntax Integration Methods.}
Previously we compare the performances of several syntax-based models, e.g.,
ASGCN \cite{zhang-etal-2019-aspect},
TD-GAT \cite{huang-carley-2019-syntax},
RGAT \cite{wang-etal-2020-relational},
and our model.
In addition, we take into consideration other types of state-of-the-art syntax-aware ABSA models in recent works, such as DGEDT \cite{tang-etal-2020-dependency}, KumaGCN \cite{chen-etal-2020-inducing}.
Note that both ASGCN and our model use GCN to encode syntactic dependency structure, while our model additionally navigates the syntax labels and aspect into the modeling.
TD-GAT employs a graph attention network (GAT) \cite{VelickovicCCRLB18} to encode dependency tree.
RGAT reshapes the original syntax tree into new one rooted at a target aspect.
Besides of encoding the dependency tree, DGEDT additionally considers the flat representations learnt from Transformer, while KumaGCN leverages the latent syntax structure.
We also implement an ABSA model encoding random tree for comparison.

We measure their performances\footnote{
We normalize each value by dividing the max one on each sub set.
} on five subsets of robustness testing (\emph{Restaurant}), as plotted in Fig. \ref{radar}.
We obtain some interesting patterns, that different models have distinct capabilities on each type of robustness test.
Among all the kind, our USGCN-based system performs the best on {\normalsize{R}\scriptsize{EV}\normalsize{T}\scriptsize{GT}}, {\normalsize{A}\scriptsize{DD}\normalsize{D}\scriptsize{IFF}} and {\normalsize{M}\scriptsize{A}\normalsize{M}\scriptsize{S}} challenges.
And RGAT gives the strongest performance on {\normalsize{R}\scriptsize{EV}\normalsize{N}\scriptsize{ON}} test, while DGEDT is most reliable on {\normalsize{R}\scriptsize{WT}\normalsize{B}\scriptsize{G}} test.
Also we note that encoding random trees gives the worst results on all attributes.
And encoding the latent structure (KumaGCN) actually helps little with robustness, largely due to the noise introduction.

\begin{figure}[!t]
\centering
\includegraphics[width=0.75\columnwidth]{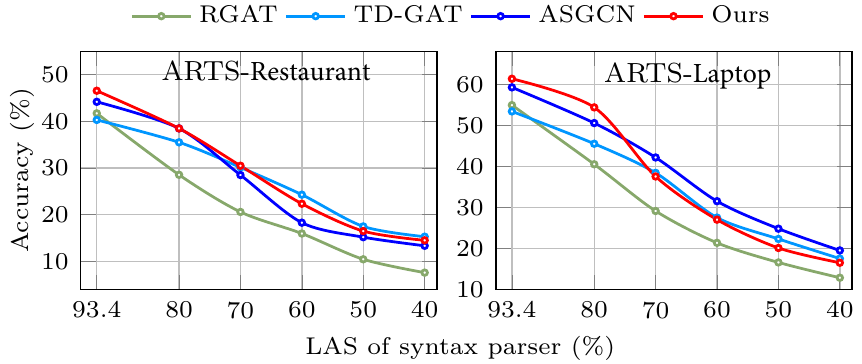}
% \vspace{-15pt}
\caption{
Influence of the syntax quality.
}
\label{synQua}
% \vspace{-10pt}
\end{figure}

\subsubsection{Faithfulness of Opinion Clues of Aspect.}
The key to robust ABSA model (for both the `Aspect-context binding' and `Multi-aspect anti-interference' challenge) lies in the capability of locating the exact opinion texts for the target aspect, i.e., the faithfulness of target aspect's opinion clues.
To confirm this faithfulness, we experiment with the manual TOWE test set \cite{fan-etal-2019-target} where the exact opinion expressions of each target aspect are explicitly annotated.
We measure the deviation between the highly weighted words decided by a ABSA model and the gold opinion expressions, which is taken as the faithfulness.
We make comparisons between different syntax-aware models, additionally including the attention-based AttLSTM model.
In Fig. \ref{violinplot} we plot the results.
We clearly see that different models come with varying faithfulness.
For example, RGAT, ASGCN and our USGCN-based model gives much lower deviations than other models.
And all the syntax-aware models show higher faithfulness than AttLSTM.

\subsubsection{Impacts of Syntax Quality.}
The quality of the syntax is crucial to the syntax-based models, since it influences the performances of robustness test.
However, ABSA data has no gold syntactic dependency annotations, and therefore we take the automatic parses instead.
By controlling the quality of the dependency parser, i.e., having varying testing LAS, we obtain an array of parsers with different quality. 
We use these parsers to general annotations in varying quality. 
We then perform the experiment and observe the corresponding performances.
Fig. \ref{synQua} shows the robustness testing accuracy under varying quality of parser.
We see that with the decreasing of parse quality, the performance drops dramatically.
Interestingly, the RGAT model performs the worst when the syntax quality decreases, mostly because reshaping the suboptimal syntax structure will dramatically introduce noises.
Besides, comparing with ASGCN, our model is more sensitive on the quality, as it additionally relies on the syntax label information.

\begin{figure}[!t]
\centering
\includegraphics[width=0.75\columnwidth]{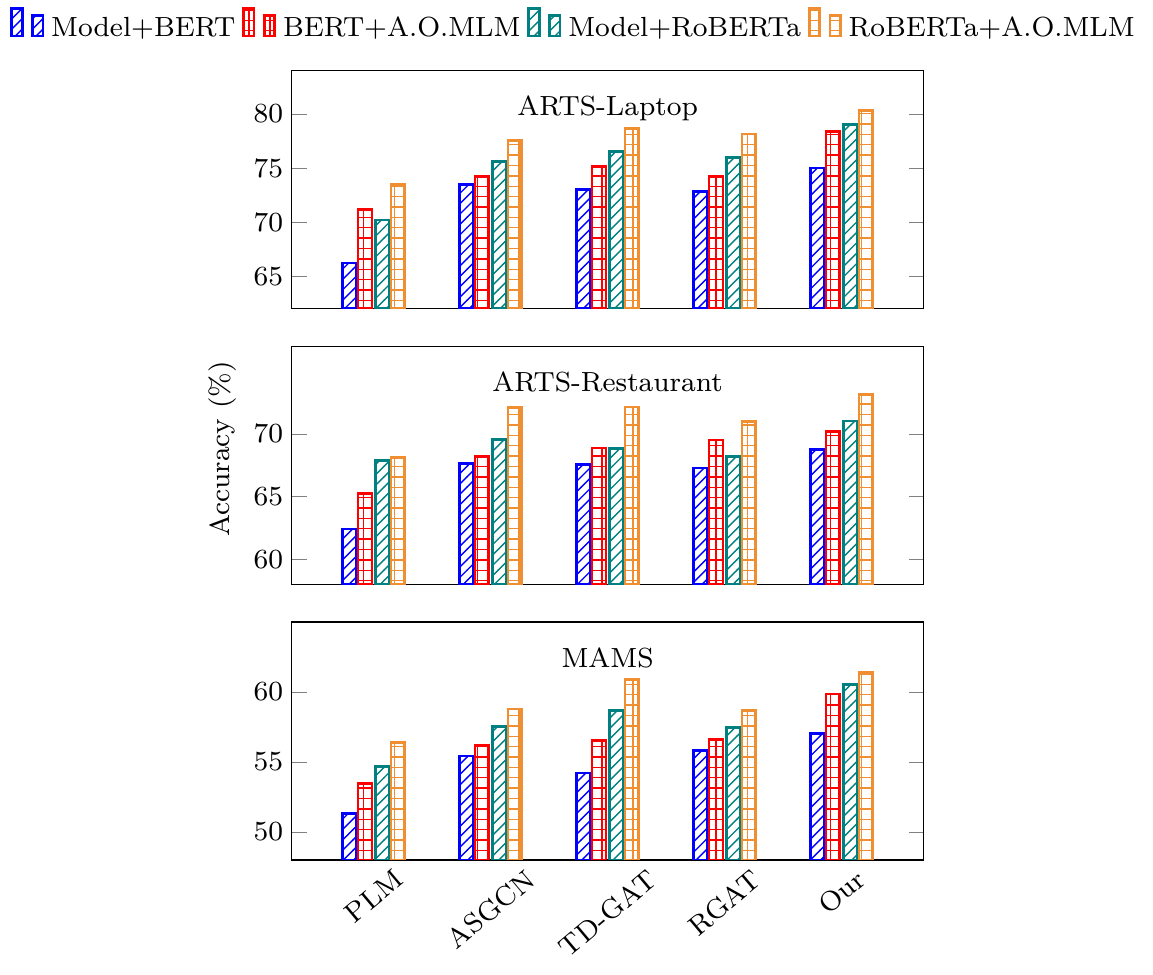}
% \vspace{-10pt}
\caption{
Performances with different pre-trained language models.
}
\label{PLM}
% \vspace{-10pt}
\end{figure}

\subsubsection{Effect of Pre-trained Language Model.}
The robustness of ABSA models are universally improved from BERT in that PLM entailed abundant linguistic and semantic knowledge for reasoning the relation between aspect and valid contexts, which coincides with related works \cite{Xu19Failure,jiang-etal-2019-challenge,xing-etal-2020-tasty}.
Here we try to explore if we can obtain better results with enhanced PLMs, e.g., other type of PLM, task-aware pre-training.
First, we compare BERT with RoBERTa\footnote{\url{https://github.com/pytorch/fairseq/tree/master/examples/roberta}}, an upgraded version of BERT.
Besides, we additionally perform a `post-training' of PLMs between the pre-training and fine-tuning stages, i.e., based on the synthetic data ($\mathbb{D}_a$) predicting the opinion texts of a given aspect via masked language modeling (MLM) technique (`A.O.MLM').
From the trends in Fig \ref{PLM} we see that comparing to BERT, RoBERTa has been shown to give very substantial improvements.
Further with a post-training of aspect-opinion MLM, each BERT/RoBERTa-wise model obtains improved results prominently.

\subsection{Corpus Evaluation}

We study two major questions w.r.t the synthetic data induction.

% \vspace{-7pt}
\begin{itemize}
    \item[\textbf{Q1}:] \emph{What is the contribution from each type of three different pseudo data?}
    \item[\textbf{Q2}:] \emph{How does the quality of pseudo data influence the robustness learning?}
\end{itemize}
% \vspace{-8pt}

\subsubsection{Contributions from Different Type of Synthetic Data.}
\begin{figure*}[!t]
\centering
\includegraphics[width=0.9\textwidth]{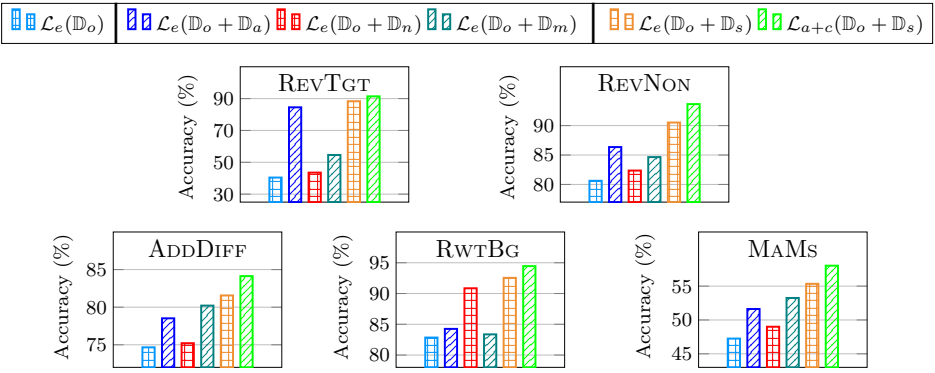}
% \vspace{-10pt}
\caption{
Training using additional synthetic data of different types, and evaluating on each specific robustness test set.
}
\label{Data-subset}
% \vspace{-8pt}
\end{figure*}

Each type of our constructed pseudo training data ($\mathbb{D}_a, \mathbb{D}_n$ and $\mathbb{D}_m$) is devoted to enhancing the robust challenges from different perspectives.
Here we examine the contribution of each type of the data to different robust testing subsets.
In Fig. \ref{Data-subset} we show the results (based on \emph{Restaurant}), from which we gain some interesting observations.
First of all, it is clear that the \emph{sentiment modification} data ($\mathbb{D}_a$) contributes the most to {\normalsize{R}\scriptsize{EV}\normalsize{T}\scriptsize{GT}} and {\normalsize{R}\scriptsize{EV}\normalsize{N}\scriptsize{ON}}, where the former takes the major proportion in the overall robustness test.
This is reasonable since enriching the sentiment diversification of each target aspect with various opinion words via $\mathbb{D}_a$ can directly enhance the capability of the first \textbf{\em aspect-context binding} challenge, making the ABSA model more correctly linking the target aspects to the critical opinion clues.

Second, the \emph{non-target aspects addition} data ($\mathbb{D}_m$) more benefits {\normalsize{A}\scriptsize{DD}\normalsize{D}\scriptsize{IFF}} and {\normalsize{M}\scriptsize{A}\normalsize{M}\scriptsize{S}}, while the \emph{background rewriting} data ($\mathbb{D}_n$) mainly improves {\normalsize{R}\scriptsize{WT}\normalsize{B}\scriptsize{G}}.
It is also easy to understand.
Because we in $\mathbb{D}_m$ increase the number of non-target aspects in sentences, creating the rich cases of multiple aspect coexistence for facilitating the learning of ABSA model.
When facing with the multi-aspect challenge in {\normalsize{A}\scriptsize{DD}\normalsize{D}\scriptsize{IFF}} and {\normalsize{M}\scriptsize{A}\normalsize{M}\scriptsize{S}}, the model naturally gives better performances.
Finally, when combining the full set of all three data ($\mathbb{D}_o+\mathbb{D}_s$) all the robust challenges receive the highest results, which, notably, can be further enhanced by using better training strategies, i.e., $\mathcal{L}_{a+c}(\mathbb{D}_o+\mathbb{D}_s)$).
Also we notice that any use of our enhanced data will improve the robustness, in the comparison with the setting of $\mathcal{L}_{e}(\mathbb{D}_o)$).

\subsubsection{Impacts of the Pseudo Data Quality.}
\begin{figure*}[!t]
\centering
\includegraphics[width=0.83\textwidth]{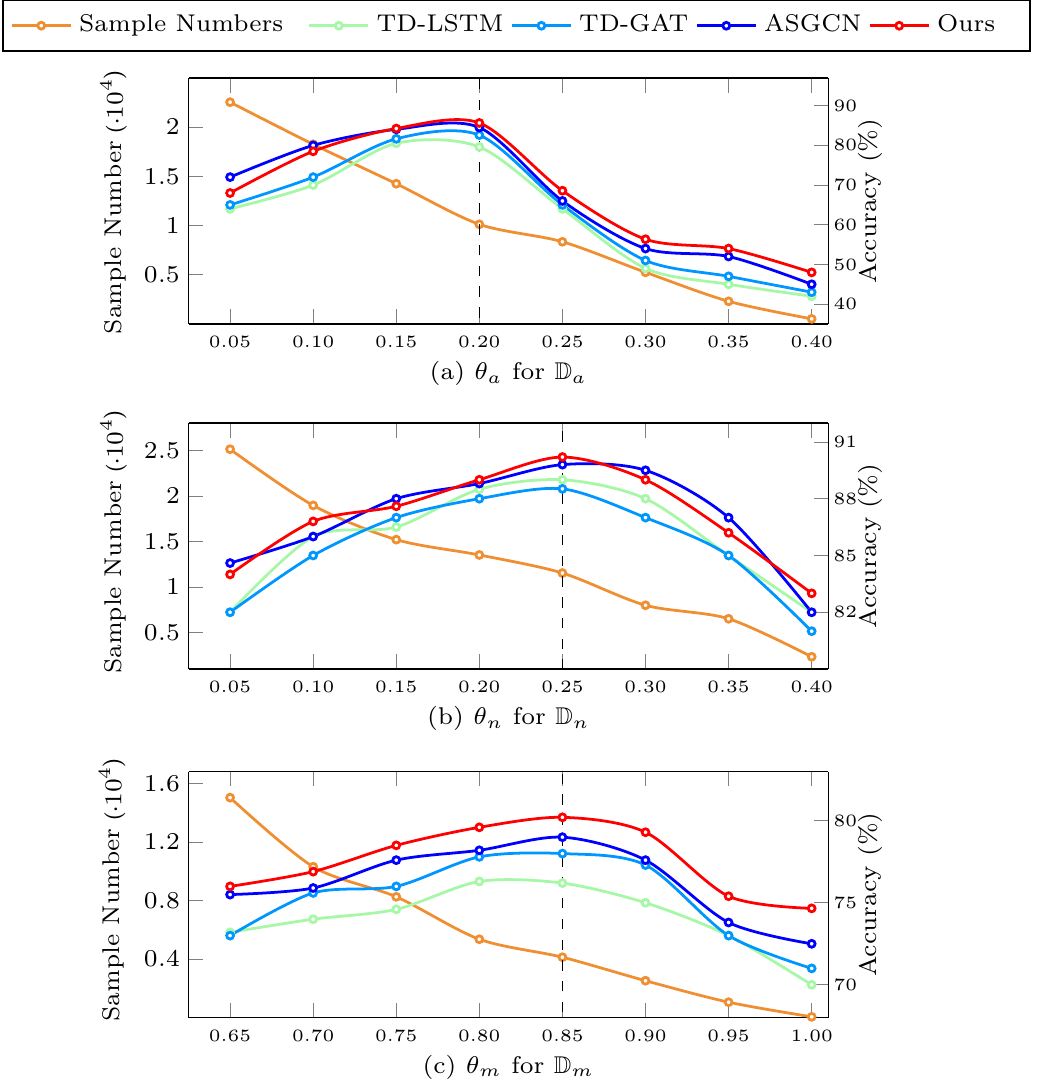}
% \vspace{-13pt}
\caption{
Results under different quality of synthetic data.
}
\label{Data-qua}
% \vspace{-10pt}
\end{figure*}

In $\S$\ref{Synthetic Corpus Construction} we devise three threshold values respectively, i.e., $\theta_a, \theta_n$ and $\theta_m$, for the quality control of each corresponding data.
Now we study the influences of the constructed synthetic data quality.
Fig. \ref{Data-qua} plots the curves of the performances under varying threshold values.
First, we see that with the increasing of the thresholds (inclusively $\theta_a, \theta_n$ and $\theta_m$), all the numbers of induced samples are reduced dramatically.
Intuitively, these constructed samples with higher qualities are always the minority.
On the other hand, ABSA models achieve their best performances in the trade-off between data quantity and quality.
In other words, too few number of training data causes insufficient signals to learn the inductive bias, though with comparatively high-quality of training instances.
However, large number of training data with noisy signals also undermines the learning.
The equilibrium points vary among different types of the synthetic data, e.g., $\theta_a$=0.2, $\theta_n$=0.25 and $\theta_m$=0.85.
At the same time, the sample numbers in $\mathbb{D}_a$, $\mathbb{D}_n$ and $\mathbb{D}_m$ are 10,000, 12,500 and 4,000 approximately.
In other perspective, we find that our USGCN model consistently performs the best in any case among three data sets.

\subsection{Training Evaluation}
We have confirmed earlier that better training strategies further help improve robustness.
Correspondingly, we care about one main questions:
% \vspace{-7pt}
\begin{itemize}
    \item[\textbf{Q1}:] \emph{How does the training paradigm affect the robustness learning?}
    % \item[\textbf{Q2}:] \emph{How varying are the performances of each contrastive learning scheme?}
\end{itemize}
% \vspace{-8pt}

\noindent In $\S$\ref{Contrastive Learning} we propose total four types of learning paradigms to fully utilize the rich contrastive signals within the synthetic corpus unsupervisedly.
Furthermore, we now try to explore:
% \vspace{-7pt}
\begin{itemize}
% \setlength{\topsep}{0pt}
% \setlength{\itemsep}{0pt}
% \setlength{\parsep}{0pt}
% \setlength{\parskip}{0pt}
    % \item[\textbf{Q1}:] \emph{How does training paradigm affect the robustness learning?}
    \item[\textbf{Q2}:] \emph{How varied are the performances of each contrastive learning scheme?}
\end{itemize}
% \vspace{-8pt}

\subsubsection{Visualization for Advanced Training strategy.}
\begin{figure}[!t]
\centering
\includegraphics[width=0.96\columnwidth]{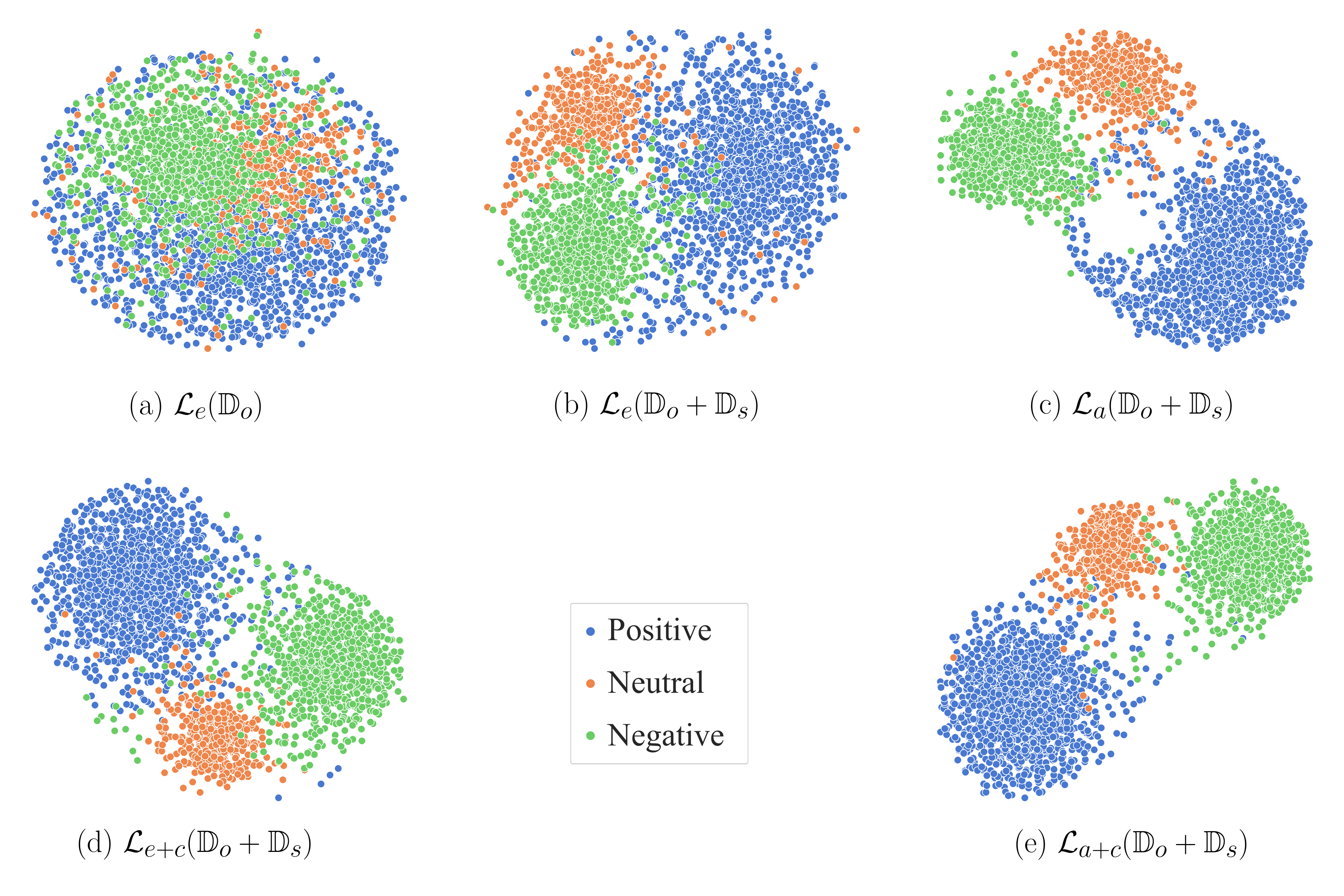}
% \vspace{-5pt}
\caption{
Visualizations of the model representations by different training strategies.
Best viewed in color and by zooming in.
}
\label{visual}
% \vspace{-10pt}
\end{figure}

\textbf{Q1} asks for the underlying reason that different training methods lead to diversified performances.
As we introduced earlier, the adversarial training help to reinforce the perception of contextual change in the help with three type of enhanced pseudo data, while the contrastive learning can unsupervisedly consolidate the recognition of different labels.
% explore
To confirm this, we consider empirically performing visualization of the resulting model representations by different training strategies, e.g., adversarial training ($\mathcal{L}_a$) and contrastive learning ($\mathcal{L}_c$) as well as the hybrid training ($\mathcal{L}_{e+c}$ and $\mathcal{L}_{a+c}$).
We render the final feature representation $\bm{r}^f$ of each instance in {\normalsize{A}\scriptsize{RTS}} test set (\emph{Restaurant}) with T-SNE algorithm, as shown in Fig. \ref{visual}.
It is quite easy to see the gaps of the models' capability between each training method.
First of all, from the patterns between (b) and (c) we understand that the decision boundaries learnt by the standard cross-entropy training objective can be quite obscure, while the advanced training, especially the adversarial training, indeed helps greatly in learning clearer decision boundaries between different sentiment labels.
Besides, without employing the adversarial training, instead we ensemble the cross-entropy training objective with additional unsupervisedly contrastive representation learning, the decision boundaries can also became much clearer.
This reflects the importance of leveraging contrastive representation learning for sufficiently mining the inherent knowledge in the data for better ABSA robustness.
Also notedly, we see that the hybrid training of adversarial training and contrastive learning ($\mathcal{L}_{a+c}$) helps to give the best effect.
Additionally, comparing Fig. \ref{visual}(a) with Fig. \ref{visual}(b) we understanding the high effectiveness of leveraging the pseudo training corpus.

\begin{figure}[!t]
\centering
\includegraphics[width=0.64\columnwidth]{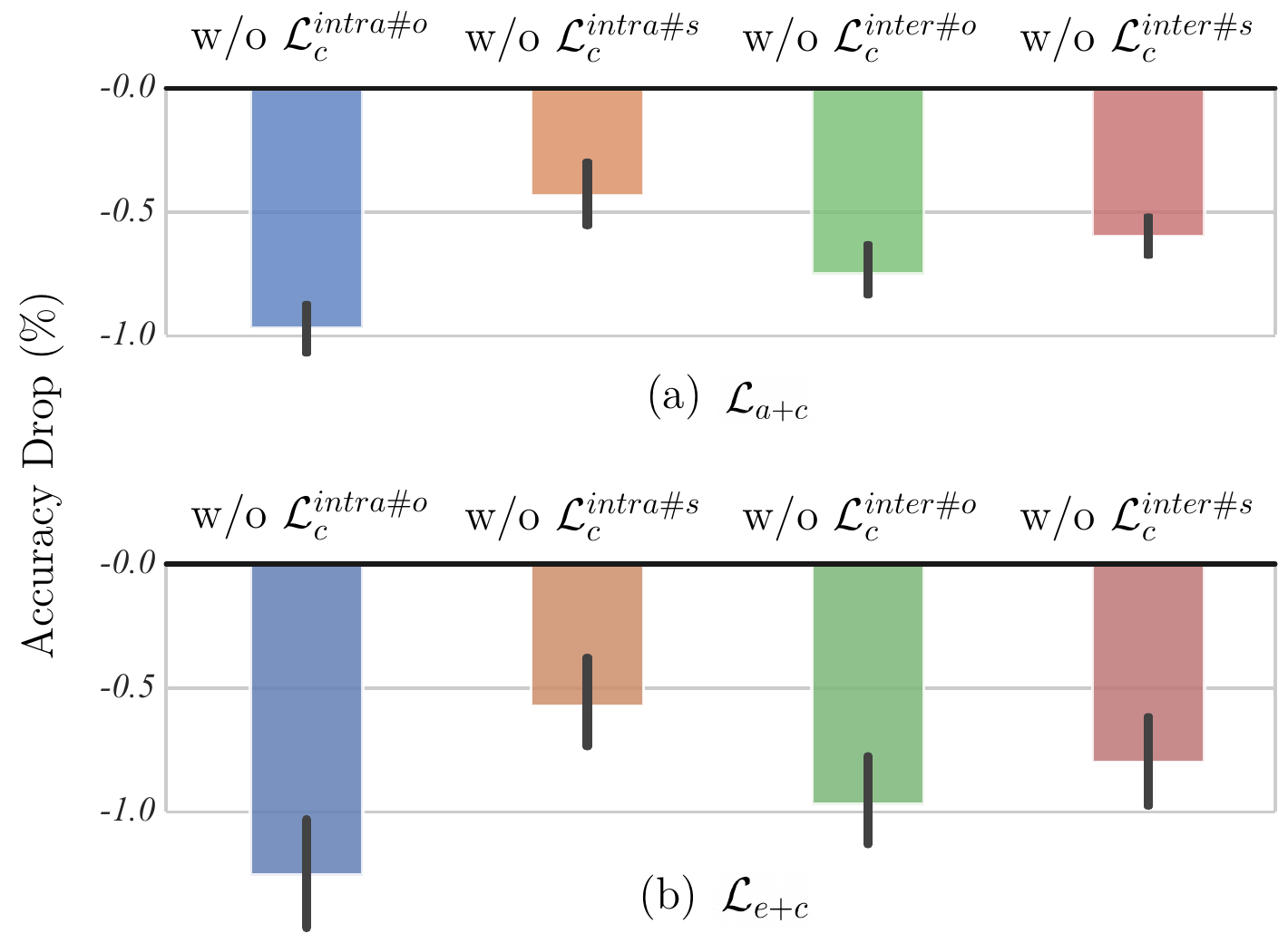}
% \vspace{-10pt}
\caption{
Performances by different contrastive learning schemes.
}
\label{CL}
% \vspace{-10pt}
\end{figure}

\subsubsection{Into the Contrastive Learning.}
Each type of the contrastive learning schemes focuses on one case of intra/inter-aspect and opinion/structure-guided perspectives.
In all the above experiments we take the total form of all these learnings in the pursuit of maximum effect.
Here we would like to check the contribution of each, separately.
We reach the goal by ablating each loss term and observing the corresponding performance drop.
Intuitively, the bigger the drop is the more important it should be.
We plot the results of accuracy drops for $\mathcal{L}_{e+c}$ and $\mathcal{L}_{a+c}$ in Fig. \ref{CL}.
In general, the drops by $\mathcal{L}_{e+c}$ are higher than that by $\mathcal{L}_{a+c}$.
The most plausible reason could be that the adversarial training alone can learn good biases, compared with the training with cross-entropy objective.
% (with $\mathcal{L}_{a+c}$$>$$\mathcal{L}_{e+c}$)
Besides, from a global view, for both $\mathcal{L}_{e+c}$ and $\mathcal{L}_{a+c}$, we witness the same trend, i.e., opinion-guided learning is primary than the structure-guided one.
This largely proves that the final feature representations at last aggregation layer carry the major opinion features for the target aspect.
In contrast, the representation from syntax fusion module at middle layer may not able to fully cover the final opinion-aware feature representation.
% most of the major effect of 
But we note that, within the scope of inter-aspect learning, the role of structure-guided one is on par with opinion-guided one.
This is because two different aspects can have clearer distinction in syntax structures, allowing for a better contrast.
Overall, all these four types of learning schemes can contribute the ABSA robustness.

\section{Conclusion and Future Work}
\label{Conclusion and Future Work}

Within the last decade, a good amount of ABSA neural models are emerged for pursuing stronger task performances and higher testing scores.
They however could be vulnerable to new cases in the wild where the contexts can be varying.
Improving the ABSA robustness thus becomes imperative.
In this study, we rethink the bottleneck of ABSA robustness, and improve it from a systematic perspective, i.e., model, data and training.
We propose improving the ABSA model robustness, strengthening the adapting capability of the model in real-world environments and facilitating the commercial applications for our society. 
Also the methods we proposed for the robustness improvements of ABSA scenario can effortlessly transfer to other AI-technique based applications and tasks, and thus benefit the society.
Following we conclude what works for robust ABSA, and then shed light on what's the next.

After a comprehensive comparison between current strong-performing ABSA models, syntax-based models show the best robustness among others, due to their extraordinary capability on locating exact opinion texts for target aspect.
In this work we introduce a novel syntax-aware model: we model the syntactic dependency structure and the arc labels as well as the target aspect simultaneously with a GCN encoder, namely universal-syntax GCN (USGCN).
With USGCN, we achieve the goal of navigating richer syntax information for best ABSA robustness.
Also we reveal that better pre-trained language models help robustness learning greatly.
As future work, we encourage to either relieve the negative effect of syntax-based methods (e.g., relying much on syntax quality) or devise syntax-agnostic models but with strong aspect-context binding abilities.
Alternatively, we recommend integrating external syntax knowledge into PLMs during the post-training stage and then performing opinion-aware fine-tuning.

Another key bottleneck is the data.
Strong ABSA models achieve good accuracy on in-house testing data but fail to scale to unseen cases, because the insufficiency of learning good inductive bias on training set.
We thus construct additional synthetic training data.
Three types of high-quality corpora are automatically induced based on raw SemEval data, enabling sufficient robust learning of ABSA models.
Each type of pseudo data aims to improve one certain angle of ABSA robustness.
Future work may explore better approaches to automatically construct higher quality of corpus, e.g., inducing more reliable data with less sentiment-uncertainty.
Besides, automatically constructing large-scale sentiment data for training better PLM for robust ABSA will be a promising direction.

The training paradigm is also important.
Most of existing ABSA frameworks take the standard training with negative cross-entropy objective.
In this work, we propose to perform adversarial training based on the pseudo data to enhance the resistance to the environment perturbation, such as opinion flip, background rewriting, and multi-aspects coexistence.
Meanwhile, we employ the unsupervised contrastive learning technique for further enhancement of representation learning, based on the contrastive samples in pseudo data.
We design four different learning schemes to fully consolidate the recognition of robustness challenges.
As future work, we believe it will be meaningful to build more reasonable and efficient adversarial training framework, achieving higher robustness performance in fewer time-cost.

\bibliographystyle{ACM-Reference-Format}
\bibliography{ref}

\end{document}